\documentclass[acmsmall]{acmart}

\usepackage{graphicx}
\usepackage{subfigure}
\usepackage{algorithm}
\usepackage{algorithmic}
\usepackage{lipsum}

\AtBeginDocument{%
	\providecommand\BibTeX{{%
			\normalfont B\kern-0.5em{\scshape i\kern-0.25em b}\kern-0.8em\TeX}}}

\begin{document}
	\newcommand{\te}{{\theta}}
	\newcommand{\ts}{{\text{s}}}
	\newcommand{\tp}{{\text{p}}}
	\newcommand{\tg}{{\text{g}}}
	
	\newcommand{\bfx}{{\mathbf{x}}}
	\newcommand{\bfh}{{\mathbf{h}}}
	\newcommand{\bfo}{{\mathbf{o}}}
	\newcommand{\bfp}{{\mathbf{p}}}
	
	\newcommand{\calD}{{\mathcal{D}}}
	\newcommand{\calDk}{{\mathcal{D}^k}}
	\newcommand{\calP}{{\mathcal{P}}}
	\newcommand{\calPk}{{\mathcal{P}^k}}
	
	\newcommand{\calL}{{\mathcal{L}}}
	\newcommand{\calLk}{{\mathcal{L}^k}}
	
	\newcommand{\calLbf}{{\mathcal{L}_{\text{Bf}}}}
	\newcommand{\calLkbf}{{\mathcal{L}^k_{\text{Bf}}}}
	
	\newcommand{\Dk}{{D^k}}
	\newcommand{\Dte}{{D_{\text{te}}}}
	\newcommand{\Dktr}{{D^k_{\text{tr}}}}
	\newcommand{\Dkte}{{D^k_{\text{te}}}}
	
	\newcommand{\Ntr}{{N_{\text{tr}}}}
	\newcommand{\Nte}{{N_{\text{te}}}}
	\newcommand{\Nktr}{{N^k_{\text{tr}}}}
	\newcommand{\Nkte}{{N^k_{\text{te}}}}
	
	\newcommand\blfootnote[1]{%
		\begingroup
		\renewcommand\thefootnote{}\footnote{#1}%
		\addtocounter{footnote}{-1}%
		\endgroup
	}

	\title{Aggregate or Not? Exploring Where to Privatize in DNN Based Federated Learning Under Different Non-IID Scenes}
	
	% \blfootnote{This research was supported by National Natural Science Foundation of China (Grant Nos. 61632004, 61773198, and 41901270), NSFC-NRF Joint Research Project under Grant 61861146001, and Natural Science Foundation of Jiangsu Province (Grant No. BK20190296).}
	
	\blfootnote{This is generated with the ACM Journal LeTex format.}
	
	\author{Xin-Chun Li}
	\email{lixc@lamda.nju.edu.cn}
	\author{Le Gan}
	\email{ganle@nju.edu.cn}
	
	\author{De-Chuan Zhan$^*$}\blfootnote{*Corresponding Author}
	\email{zhandc@nju.edu.cn}
	\affiliation{%
		\institution{State Key Laboratory for Novel Software Technology, Nanjing University}
	}

	\author{Yunfeng Shao}
	\email{shaoyunfeng@huawei.com}
	
	\author{Bingshuai Li}
	\email{libingshuai@huawei.com}
	
	\author{Shaoming Song}
	\email{shaoming.song@huawei.com}
	
	\affiliation{%
		\institution{Huawei Noah's Ark Lab}
	}
	\renewcommand{\shortauthors}{Li, Gan and Zhan, et al.}
	
	\begin{abstract}
		Although federated learning (FL) has recently been proposed for efficient distributed training and data privacy protection, it still encounters many obstacles. One of these is the naturally existing statistical heterogeneity among clients, making local data distributions non independently and identically distributed (i.e., non-iid), which poses challenges for model aggregation and personalization. For FL with a deep neural network (DNN), privatizing some layers is a simple yet effective solution for non-iid problems. However, which layers should we privatize to facilitate the learning process? Do different categories of non-iid scenes have preferred privatization ways? Can we automatically learn the most appropriate privatization way during FL? In this paper, we answer these questions via abundant experimental studies on several FL benchmarks. First, we present the detailed statistics of these benchmarks and categorize them into covariate and label shift non-iid scenes. Then, we investigate both coarse-grained and fine-grained network splits and explore whether the preferred privatization ways have any potential relations to the specific category of a non-iid scene. Our findings are exciting, e.g., privatizing the base layers could boost the performances even in label shift non-iid scenes, which are inconsistent with some natural conjectures. We also find that none of these privatization ways could improve the performances on the Shakespeare benchmark, and we guess that Shakespeare may not be a seriously non-iid scene. Finally, we propose several approaches to automatically learn where to aggregate via cross-stitch, soft attention, and hard selection. We advocate the proposed methods could serve as a preliminary try to explore where to privatize for a novel non-iid scene.
		
		% The learned privatization ways lead to soft combinations between shared and private components and inspire novel privatization ideas.
	\end{abstract}
	
	
	

	\maketitle

	\section{Introduction}
	Although traditional machine learning methods have experienced great success in some fields~\cite{AlexNet,ResNet}, the samples must be centralized for training. However, due to the data privacy or transmission cost in real-world applications, data from individual participants are often isolated and can not be located on the same device~\cite{Gboard}. Derived but different from the standard distributed optimization~\cite{LargeScaleNet}, federated learning (FL)~\cite{Fed-Advances,Fed-Concept,FedAvg} is tailored for data privacy protection and efficient distributed training. FL has been applied to many specific tasks, e.g., language modeling~\cite{Federated-ngram}, medical relation extraction~\cite{Fed-MedicalRE}, industrial topic modeling~\cite{Federated-ngram}, and medical image analysis~\cite{FLOP-Medical}. 
	
	Standard FL methods utilize a parameter server~\cite{ParameterServer} to collaborate isolated clients, dividing the distributed training process into multiple rounds of local and global procedures. During the local procedure, a fraction of clients is selected to tune the global model downloaded from the parameter server. During the global procedure, the server collects the model updates from clients and fuses them into the global model. Because local clients bear more computation, FL can reduce the communication cost to some extent. From another aspect, only model parameters are transmitted among clients and the server, and hence FL provides essential privacy protection for local data.
	
	FL also faces many challenges. First, a real-world FL system may contain millions of local clients (e.g., portable devices), in which the communication may be unstable and the computation budget of local devices is often limited. This obstacle is categorized as the systematical heterogeneity in~\cite{FedProx}, and the possible solutions are investigated in~\cite{Fed-System,Fed-strategy-communication}. Second, although FL avoids direct transmission of users' data and provides a certain level of privacy protection for users, it may encounter attacks or model inspection from adversaries in non-ideal real-world applications~\cite{Fed-Advances}. Some advanced privacy techniques can be applied to FL to acquire stricter privacy requirements, e.g., differential privacy~\cite{DP-Algo,FedDP-Algo}, homomorphic encryption~\cite{FedHE}. Third, local clients' various contexts determine that their data are non independently and identically distributed (i.e., non-iid). This problem is categorized as statistical heterogeneity in~\cite{FedProx}. As the most standard FL algorithm, FedAvg~\cite{FedAvg} is originally declared robust to the non-iid data, while later studies have shown that it could experience severe performance degradation when faced with heterogeneous data~\cite{Fed-NonIID-Quagmire}. One convincing explanation is the weight divergence proposed in~\cite{Fed-NonIID-Data}, pointing out that the local update could diverge significantly from the optimal global target. Solutions with various considerations are proposed to enhance the model performances in FL with non-iid data~\cite{FedCurv,FedFusion,Scaffold,HMR,Fed-MAML,PFL-DA-CoRR2019}. In this paper, we majorly investigate the non-iid challenge in FL.
	
	There are two specific goals when considering boosting the FL performances, i.e., aggregation and personalization. The goal of aggregation is to obtain a complete global shared model applicable to all participating clients and expect it could be quickly adapted to forthcoming clients. FedAvg~\cite{FedAvg} is primitively proposed to aggregate a well-performed global model with the collaboration of local clients. As aforementioned, the statistical heterogeneity leads to diverged local model updates, which poses a huge challenge for model aggregation. Several solutions have been proposed to tackle this problem. Some methods~\cite{Fed-Unbias,HMR,Fed-FD-FAug} introduce a small public shared dataset on the parameter server and utilize it to fine-tune the aggregated model. Some other methods~\cite{FedMD,FedEnsembleDistll} take advantage of knowledge distillation~\cite{KD} to fuse the local models' ability into the global model. These methods require an appropriate global dataset on the server, which is not easy to collect with stricter privacy constraints. Limiting the scope of model updates during the local procedure is another way to solve this problem. FedProx~\cite{FedProx} directly restricts the local parameters' L2 distance to the global ones; FedCurv~\cite{FedCurv} additionally introduces the fisher information to assign weights to different parameters for overcoming catastrophic forgetting during the local procedure; FedFusion~\cite{FedFusion} and FedMMD~\cite{FedMMD} impose such restrictions on the level of intermediate features; Scaffold~\cite{Scaffold} introduces control variates and momentum for stable model updates.
	
	Unlike the goal of aggregation, personalization indicates the primary incentive of local clients to participate in FL. Usually, local clients have insufficient labeled data to train well-performed models, and hence they are inclined to take part in the FL process. However, when faced with non-iid data, some clients may gain no benefit as shown in~\cite{FedAdapt,Per-Survey}. The main reason is mathematically proposed in~\cite{FedBoost}, which states that the optimal solution of federated optimization is a convex combination of local ideal hypotheses. This implies that the single global model obtained by standard FL algorithms is a compromise solution for all local clients, and naturally, it is not the optimal model for some local clients. Some researches focus on facilitating personalization in FL. Formulating the FL as a multi-task learning~\cite{Fed-MultiTask} or meta-learning~\cite{Fed-MAML} task can improve the personalized models' performances. Varying the hyper-parameter settings during the local procedure can also facilitate fine-tuning of the global model~\cite{EvalPer}. In DNN based FL, privatizing some network layers is a simple yet effective solution for better personalization~\cite{FedPer-CoRR2019,LG-FedAvg-CoRR2020,PFL-DA-CoRR2019,FURL-CoRR2019,FedDML}.
	
	This paper mainly focus on investigating private-shared models' effectiveness for both FL aggregation and personalization. This investigation's primary motivation originates from an interesting finding that existing FL methods with private-shared models take various privatization ways, e.g., privatizing the task-specific layers~\cite{FedPer-CoRR2019}, privatizing the base layers~\cite{LG-FedAvg-CoRR2020}, privatizing the user embedding layer~\cite{FURL-CoRR2019}, or even privatizing a complete deep model~\cite{PFL-DA-CoRR2019,FedDML}. Hence, we aim to investigate whether these architecture designs conform to some definite rules and which layers we should privatize when given a novel FL non-iid scene. Specifically, our investigation steps and contributions are listed as follows:
	
	\begin{itemize}
		\item {\bf We select several classical FL benchmarks and categorize them into covariate or label shift non-iid scenes}. Inspired by the categorization of non-iid scenes in~\cite{Fed-Advances}, we select four FL benchmarks that are commonly utilized in previous works, i.e., FeCifar10, FeCifar100, Shakespeare, and FeMnist. The first two benchmarks are label shift non-iid scenes where the heterogeneity is mainly caused by splitting data according to labels~\cite{Fed-NonIID-Data,Fed-NonIID-Quagmire}. The second two are benchmarks recommended by LEAF~\cite{LEAF}, and we view them as covariate shift non-iid scenes, where the heterogeneity is mainly caused by different play roles or users.
		\item {\bf We compare aggregation and personalization performances of various privatization ways based on both coarse-grained and fine-grained network splits.} For each benchmark, we first split its corresponding DNN into two coarse parts, i.e., the feature extractor (a.k.a., encoder) and the task-specific predictor (a.k.a., classifier). We construct various private-shared models and compare their aggregation and personalization performances in FL. Then, we split the DNN into fine-grained blocks for deeper analysis.
		\item {\bf We propose several automatical methods to learn where to aggregate.} To automatically learn an appropriate privatization way for a specific FL non-iid scene, we keep both shared and private models on local clients and utilize three techniques, including cross-stitch, soft attention, and hard selection, to fuse features and outputs during the FL dynamically.
	\end{itemize}
	
	We observe several exciting phenomena, e.g., privatizing the task-specific classifier may not be the most appropriate way for label shift non-iid scenes (see in Sect.~\ref{sect-coarse-privatization}); none of the investigated privatization ways improve the performances on Shakespeare (see in Sect.~\ref{sect-coarse-privatization} and Sect.~\ref{sect-auto-learn}); the personalization performances will degrade monotonically with the number of private layers becoming larger in the single branch privatization ways (see in Sect.~\ref{sect-fine-privatization}). Aside from the three automatical approaches, we also propose an interpolation trick to quickly search for the appropriate way to privatize (see in Sect.~\ref{sect-auto-learn}).

	\begin{figure}[htbp]
		\centering
		\includegraphics[width=\linewidth]{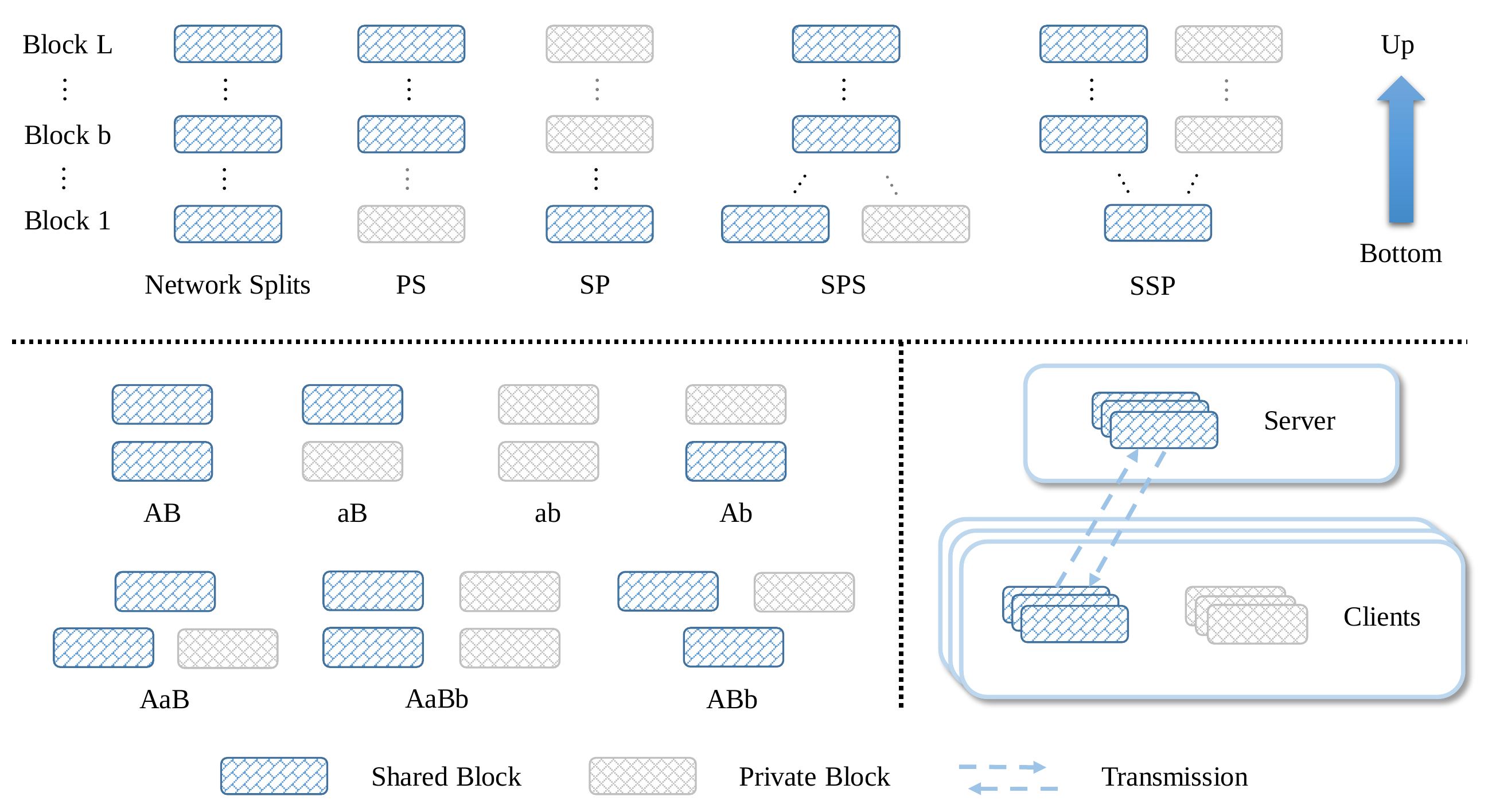}
		\centering
		\caption{{\small Illustrations of the defined privatization ways and corresponding FL framework. The top part shows the concept of ``block" and ``network splits" defined in Def.~\ref{def-block} and Def.~\ref{def-network-split}; then the defined four ``privatization" ways in Def.~\ref{def-privatization}, i.e., PS, SP, SPS, and SSP, are correspondingly presented. The bottom left part shows the unique privatization ways with $L=2$, e.g., we split a network into feature extractor (a.k.a., encoder) and task-specific predictor (a.k.a., classifier). The bottom right part shows the FL framework with private components, where only the shared blocks participate in the global aggregation.}}
		\label{fig-privatize-teaser}
	\end{figure}

	\section{Related Works}
	The most relevant learning paradigms to FL faced with non-iid data are transfer learning (TL)~\cite{Survey-Transfer} and multi-task learning (MTL)~\cite{Survey-MultiTask,Survey-MultiTask-Deep}. Hence, we will introduce some related works, including private-shared models in TL and MTL, existing researches on where to transfer/share in TL/MTL.
	
	\subsection{Private-Shared Models in TL and MTL}
	Private-shared models aim to divide private and shared information among domains/tasks via feeding data to different network components. The shared components are commonly assumed to learn domain-invariant knowledge, and the private ones can capture task-specific information. Dating back to MTL without deep networks, MTFL~\cite{MTFL} proposes to share the feature transformation matrix among tasks and learn task-specific weights for each task individually. This evolves into sharing the same feature extractor among tasks and privatizing task-specific layers in ~\cite{CNN-MTL-ICML2008,RNN-PS-IJCAI2016}. DSN~\cite{DSN-NeurIPS2016} designs a separation framework in FL with shared and private encoders for source and target domains. As a special kind of private-shared architecture, domain-specific batch normalization~\cite{DSBN} is proposed for better adaptation in TL.
	
	Similarly, private-shared models have also been applied to FL. FedPer~\cite{FedPer-CoRR2019} privatizes the task-specific classifier for each client for better personalization, while on the contrary, FedL2G~\cite{LG-FedAvg-CoRR2020} keeps the feature extractor private to process heterogeneous inputs and shares task-specific layers. FLDA~\cite{PFL-DA-CoRR2019} keeps a whole model private to mitigate the performance degradation induced by differential privacy. FedDML~\cite{FedDML} also privatizes the whole model, while they introduce knowledge distillation for better aggregation and personalization simultaneously. As a particular case, FURL~\cite{FURL-CoRR2019} advocates privatizing the user embedding layer for better federated recommendation.

	\subsection{Where to Transfer/Share in TL/MTL}
	For TL, when, what, and how to transfer are three core problems to be addressed as proposed in~\cite{Survey-Transfer}. Likewise, when, what, and how to share in MTL are corresponding fundamental issues to be solved~\cite{Survey-MultiTask}. Although the private-shared models described above provide essential solutions for dataset shift~\cite{DatasetShift} in FL and MTL, which layers should we share or privatize is still an open problem. Existing researches dive into this from both qualitative and quantitative perspectives. Qualitative ones take abundant experimental studies to explore the underlying rules. In DNN based tasks, the base layers are assumed to be more transferable than top layers, which has been verified by existing studies~\cite{NN-Transferability}. Based on this, the layer-wise transferability of deep representations is further explored in~\cite{GrandualTL}, and this study also points that the transferability is non-symmetric. Additionally, L2Tw~\cite{L2Tw} introduces meta-networks to automatically decide which features should be transferred to which target layers. The task relations are also experimentally explored in MTL~\cite{Taskonomy,NLP-Transferability}, exploring which source task is the most adaptable to the target one. There are also some quantitative works to explore the transferability among tasks, e.g., conditional entropy~\cite{ConditionalEntropy}, H-score~\cite{Hscore}, and LEEP~\cite{LEEP}.
	
	However, the problem of where to aggregate in FL under non-iid scenes has not been explicitly studied as far as we know. Therefore, we aim to qualitatively explore the relationship between specific privatization ways and FL performances under different non-iid scenes in this paper.

	\section{Background}
	This section will introduce the basic procedure of FL, the goals of model aggregation and personalization, and the specific issues that we want to explore.
	
	\subsection{Basic Procedure of FL}
	In standard FL systems, e.g., FedAvg~\cite{FedAvg}, we have $K$ distributed clients and a global parameter server. Each client has a local data distribution $\calDk=\calPk(\bfx, y)$, and $\calDk$ could differ from other local clients' distributions a lot. For the $k$th client, its optimization target is:
	\begin{equation}
		\calLk(\te^k) \triangleq E_{(\bfx^k, y^k) \sim \calDk}\left[ \ell\left( f(\bfx^k;\te^k), y^k \right) \right], \label{eq:local-target}
	\end{equation}
	where $f(\cdot;\te^k)$ is the prediction function with parameters $\theta^k$ and $f(\bfx^k;\te^k)$ is a $C$-dimensional probability vector that reflects the prediction result. $\ell\left(\cdot, \cdot\right)$ is the loss function, which is usually the cross-entropy loss. Commonly, the client $k$ could only observe a small training dataset $\Dktr=\{(\bfx_i^k, y_i^k)\}_{i=1}^{\Nktr}$ from the data distribution $\calDk$, where $\Nktr$ is the number of training instances and $i$ is the sample index. A small $\Nktr$ may limit the client's ability to train a well-performed model. Therefore, collaborating all clients is a possible solution. FedAvg takes the following optimization target:
	\begin{equation}
		\calL(\te) \triangleq \sum_{k=1}^K p_k \calL^k(\te) = \sum_{k=1}^K p_k \left[ E_{(\bfx^k, y^k) \sim \calDk}\left[ \ell\left( f(\bfx^k;\te), y^k \right) \right] \right], \label{eq:global-target}
	\end{equation}
	where $p_k$ is often set as $\Nktr / \sum_k \Nktr$ or a uniform weight $1/K$ if the clients own nearly the same amount of training samples. We take the latter one in this paper. This optimization target can be solved by rounds of local and global procedures. In the $t$th round, where $t = 1, 2, \cdots, T$, we denote $\te_{t}$ as the global parameters on the server, and $\te_{t}^k$ as the parameters on the $k$th client. Without additional declaration, we use the superscript ``$k$" to index the client and ``$i$" to index the instance; we denote notations without superscript ``$k$" as the server ones.
	
	During the local procedure in $t$th round, a subset of clients $S_t$ is selected, and the selected clients download the global model, i.e., $\te_t^k \leftarrow \te_{t}$. Then, the client $k$ samples data batches from local training dataset and update $\te_t^k$ via the corresponding empirical loss of Eq.~\ref{eq:local-target}. The updated parameters, denoted as $\hat{\te}_t^k$, will be uploaded to the global server. For the server, once it has collected updated parameters from these selected clients, it averages the model parameter via $\te_{t+1} \leftarrow \sum_{k \in S_t} \frac{1}{|S_t|} \hat{\te}_{t}^k$. These two procedures will repeat $T$ rounds until the convergence.
	
	\subsection{Goals of Aggregation and Personalization}
	As in~\cite{FedBoost}, a macroscopic analysis of FL using Bregman divergence as the loss function proves that ensemble models are optimal for the standard empirical risk minimization in FL. Specifically, we use $\mathcal{H}$ to denote the hypothesis space, and its component $h \in \mathcal{H}$ is a map of the form $h: \mathcal{X} \rightarrow \Delta\mathcal{Y}$, where $\mathcal{X}$ is the input space and $\Delta \mathcal{Y}$ is the probability simplex over the label space $\mathcal{Y}$. The loss of client $k$ with Bregman divergence is defined as:
	\begin{equation}
		\calLkbf(h^k) \triangleq B_F(\calDk || h^k ) = F(\calDk) - F(h^k) - \langle \nabla F(h^k), \calDk - h^k \rangle, \label{eq:bf-local}
	\end{equation}
	where $h^k$ is the hypothesis on the $k$th client, $F$ is a strictly convex function, and $\langle \cdot, \cdot \rangle$ denotes the inner product. Then, the global target of standard FL is to minimize:
	\begin{equation}
		\calLbf(h) \triangleq \sum_{k=1}^K p_k \calLkbf(h) = \sum_{k=1}^K p_k B_F(\calDk || h ), \label{eq:bf-global}
	\end{equation}
	
	The lemma below shows the optimality in FedAvg:
	\begin{lemma}[Optimal Solution of FedAvg~\cite{FedBoost}]
		Suppose the loss is a Bregman divergence $B_F$, let $p \in \Lambda$ be fixed and $\sum_{k=1}^K p_k \calDk \in \mathcal{H}$. Then $h^*=\sum_{k=1}^K p_k \calDk$ is a minimizer of $\sum_{k=1}^K p_k B_F(\calDk||h)$. Furthermore, if $F$ is strictly convex, then the minimizer is unique.
	\end{lemma}
	
	This inspires that some standard FL algorithms only aim to obtain a tradeoff solution among heterogeneous clients, which is not the optimal solution for some local clients. An obvious fact is that the $B_F(\calDk || h^*)=B_F(\calDk || \sum_{k=1}^K p_k \calDk)$ could be very large if the local data distribution $\calDk$ is totally distinct from the ``global distribution", i.e., $\sum_{k=1}^K p_k \calDk$. This implies that the standard FL algorithms could not satisfy all users via a single global model. From another aspect, paying efforts to accomplish the aggregation goal may not necessarily meet the personalization performances required by local clients. Therefore, an effective FL method should focus on both aggregation and personalization performances. We formally present the goals and corresponding metrics of aggregation and personalization as follows:
	\begin{definition}[Goal of Aggregation in FL] \label{def-aggregation}
		Suppose we have a global test set $\Dte=\{(\bfx_i, y_i)\}_{i=1}^M$ on the parameter server. We calculate the classification accuracy of the {\em aggregated model} in the $t$th round to evaluate the quality of aggregation, i.e., $\text{Ag}_t \triangleq \frac{1}{M} \sum_{i=1}^M \mathcal{I}\{\arg\max f(\bfx_i; \te_{t+1}) = y_i\}$.
	\end{definition}
	
	\begin{definition}[Goal of Personalization in FL] \label{def-personalization}
		Suppose we have a local test set $\Dkte=\{(\bfx_i^k, y_i^k)\}_{i=1}^{\Nkte}$ on the $k$th client, then we can obtain the performance of the {\em personalized model} with parameters $\hat{\te_t^k}$ in the $t$th round, i.e., $\text{Ap}_t^k \triangleq \frac{1}{\Nkte} \sum_{i=1}^{\Nkte} \mathcal{I}\{\arg\max f(\bfx_i^k; \hat{\te}_{t}^k) = y_i^k\}$. Then we average the personalization accuracies across selected clients as the personalization performance, i.e., $\text{Ap}_t \triangleq \sum_{k \in S_t} \frac{1}{|S_t|} \text{Ap}_t^k$.
	\end{definition}
	
	$M$ is the number of global test samples, $\Nkte$ is the number of local test samples on client $k$, $\arg\max f(\cdot; \cdot)$ returns the predicted label, and $\mathcal{I}\{\cdot\}$ is the indication function. We can find that the goal of aggregation emphasizes the global {\em aggregated model}'s ability, which is expected to work well universally for existing clients or novel clients. However, the goal of personalization emphasizes the {\em personalized model}'s ability, which reflects the original incentive for local clients to participate in FL. Due to the non-iid challenge, it is hard to generate a well-performed aggregated model. Even we can obtain such a global model, it could still perform poorly on local clients, and delicate personalization methods should be further utilized for better adaptation.
	
	\subsection{Issues to Explore} \label{sect-issues}
	A simple yet effective approach to simultaneously accomplish the goals of aggregation and personalization with non-iid data is to privatize some network components. Take the privatization of a whole model for example, we keep both a shared model $h_{\ts}$ and a private one $h_{\tp}^k$ on the $k$th client. We utilize the additive model to combine them, i.e., $h^k \triangleq (1-\alpha) h_{\ts} + \alpha h_{\tp}^k$. $\alpha \in [0,1]$ is a tradeoff coefficient. We let the shared model $h_{\ts}$ participate in the aggregation process, while the private one does not. We expect $h_{\ts}$ and $h_{\tp}^k$ can capture the client-invariant and client-specific information respectively. We also propose a simple analysis via the Bregman divergence, i.e.,
	\begin{proposition}
		If $F$ is strictly convex and $B_{F}$ is jointly convex, we denote the target loss of FL with private-shared models and present its upper bound as follows:
		\begin{equation}
			\mathcal{L}_{\text{ps}}( h_{\ts}, \{h_{\tp}^k\}_{k=1}^K ) \triangleq \sum_{k=1}^K p_k B_F( \calD^k || h^k ) \leq (1-\alpha)\sum_{k=1}^K p_k B_F ( \calD^k || h_{\ts} ) + \alpha \sum_{k=1}^K p_k B_F ( \calD^k || h_{\tp}^k ), \label{eq:ps-loss}
		\end{equation}
	\end{proposition}
	
	The proof can be easily completed via the property of Bregman divergence, i.e., $B_F( \calD^k || (1 - \alpha) h_{\ts} + \alpha h_{\tp}^k ) \leq (1 - \alpha) B_F(\calD^k || h_{\ts}) + \alpha B_F(\calD^k || h_{\tp}^k)$. We take a closer look at the upper bound in Eq.~\ref{eq:ps-loss} and we can find that the first term can be seen as the goal of aggregation, while the second term can be seen as the personalization one. $\alpha \in [0,1]$ balances the learning process of aggregation and personalization. This implies that taking advantage of private models can provide effective solutions for both aggregation and personalization in FL.
	
	However, which layers should we privatize, or must we keep a complete model private to achieve better FL performances\footnote{We refer to the aggregation and personalization performances as FL performances in a unified way.}? Although some existing methods have utilized private-shared models for accomplishing distinct goals in FL~\cite{FedPer-CoRR2019,PFL-DA-CoRR2019,LG-FedAvg-CoRR2020,FURL-CoRR2019,FedDML}, they take various privatization ways and do not explicitly investigate where to aggregate in FL under different non-iid scenes. As categorized in TL, the dataset shift~\cite{DatasetShift} contains both covariate shift and label shift scenes. Similarly, as reported in the recent comprehensive survey of FL~\cite{Fed-Advances}, the non-iid scenes contain covariate shift and label shift non-iid ones. The former assumes the heterogeneity is mainly caused by the change of the feature distribution $\calP(\bfx)$, while the conditional distribution $\calP(y|\bfx)$ is nearly the same across clients; the latter assumes the label prior distribution $\calP(y)$ varies a lot across clients, while the generation process $\calP(\bfx|y)$ is nearly the same. For example, the various styles of handwriting could lead to the covariate shift; the fact that kangaroos are only in Australia or zoos could result in label shift~\cite{Fed-Advances}. According to this, we first propose two kinds of possible conjectures for where to aggregate or privatize in FL with non-iid scenes as follows:
	\begin{enumerate}
		\item {\bf Where changes, where to privatize}. If the covariate shift in a specific FL scene is more serious, we should keep a private encoder to extract client-specific feature information. On the contrary, if the label shift is more serious, privatizing a task-specific classifier is preferred.
		\item {\bf Aggregating base layers, privatizing task-specific layers}. According to the common assumption in DNN~\cite{NN-Transferability}, base layers are more transferable among tasks, and hence we prefer to share the feature extractor among clients.
	\end{enumerate}
	
	These conjectures co-exist even in the same research area. For example, some domain adaptation (DA) methods~\cite{DAN,DaNN} follow the assumption that the base layers are more transferable and share the feature extractor between domains to align the intermediate features. However, some other DA methods~\cite{ADDA,DSBN} advocate that the covariate shift is majorly caused by the feature distribution skew and do not entirely share the feature extractor between source and target domains. Since these are just conjectures, we want to verify whether these assumptions hold in FL under different non-iid scenes.

	\begin{algorithm}[tb]
		\caption{Pseudo Code of Learn to Aggregate}
		\label{algo}
		%\textbf{ServerProcedure: \qquad \qquad \qquad \qquad \qquad \qquad \qquad \qquad \qquad}
		\flushleft{\textbf{ServerProcedure:}} \\
		\begin{algorithmic}[1]
			\FOR{global round $t = 0, 1, 2, \ldots, T$}
			\STATE $S_t \leftarrow $ sample $\max(Q \cdot K, 1)$ clients
			\FOR{$k \in S_t$}
			\STATE $\hat{\te}_{E_{\ts},t}^k$, $\hat{\te}_{C_{\ts},t}^k$, $\hat{\psi}_{t}^k \leftarrow $ ClientProcedure($k$, $\te_{E_{\ts},t}$, $\te_{C_{\ts},t}$, $\psi_{t}$)
			\ENDFOR
			\STATE Aggregation via: $\te_{E_{\ts},t+1} \leftarrow \sum_{k \in S_t}\frac{1}{|S_t|} \hat{\te}_{E_{\ts},t}^k$, $\te_{C_{\ts},t+1} \leftarrow \sum_{k \in S_t}\frac{1}{|S_t|} \hat{\te}_{C_{\ts},t}^k$, $\psi_{t+1} \leftarrow \sum_{k \in S_t}\frac{1}{|S_t|} \hat{\psi}_{t}^k$
			\ENDFOR
		\end{algorithmic}
		%\textbf{ClientProcedure}($k$, $\te_{E_{\ts},t}$, $\te_{C_{\ts},t}$, $\psi_{t}$:) \qquad \qquad \qquad \qquad \qquad \qquad
		\flushleft{\textbf{ClientProcedure}($k$, $\te_{E_{\ts},t}$, $\te_{C_{\ts},t}$, $\psi_{t}$):} \\
		\begin{algorithmic}[1]
			\STATE Download global parameters: $\te_{E_{\ts},t}^k \leftarrow \te_{E_{\ts},t}$, $\te_{C_{\ts},t}^k \leftarrow \te_{C_{\ts},t}$, $\psi_{t}^k \leftarrow \psi_{t}$
			\FOR{local epoch $e = 1, 2, \ldots, E$}
			\FOR{each batch with $B$ samples from $\calD^k$}
			\STATE Obtain the extracted features via the private and shared encoder as in Eq.~\ref{eq-feat}
			\STATE Fuse the features according to Eq.~\ref{eq-cross-stitch}/ Eq.~\ref{eq-soft-attention}/Eq.~\ref{eq-hard-selection} if use AutoCS/AutoSA/AutoHS
			\STATE Obtain the predictions of the private and shared classifier as in Eq.~\ref{eq-classify}
			\STATE Fuse the predictions according to Eq.~\ref{eq-fuse-prediction}
			\STATE Calculate loss and update all learnable parameters
			\ENDFOR
			\ENDFOR
			\STATE \textbf{Return}: the updated $\hat{\te}_{E_{\ts},t}^k$, $\hat{\te}_{C_{\ts},t}^k$, $\hat{\psi}_{t}^k$
		\end{algorithmic}
	\end{algorithm}

	\section{Methods and Algorithms}
	This section will present specific privatization methods, experimental procedures, and the proposed automatical learning algorithms.
	
	\subsection{Privatization Ways}
	Given a neural network, we have amounts of privatization ways to divide components into shared and private ones. The privatization can be a single branch one, where we directly privatize some layers and aggregate the remaining ones via the server, e.g., privatizing the base layers~\cite{LG-FedAvg-CoRR2020}, privatizing the top layers~\cite{FedPer-CoRR2019}. Meanwhile, it could be a multi-branch one, where we keep the whole model shared and additionally keep several components private, e.g., privatizing a copy of the encoder~\cite{FedFusion}, privatizing another independent model~\cite{PFL-DA-CoRR2019}. To standardize and simplify the descriptions of these privatization ways, we define the following terms:
	\begin{definition}[Block] \label{def-block}
		A block is the smallest network unit that we use to construct private-shared models. It could be a single or a series of network layers. We use uppercase letters (e.g., A, B, C) to indicate shared blocks and lowercase letters (e.g., a, b, c) to indicate private ones.
	\end{definition}
	
	\begin{definition}[Network Splits] \label{def-network-split}
		We divide a complete network into several disjoint but sequentially adjacent blocks as network splits. We use $L$ to denote the number of blocks. By default, we arrange the split blocks in alphabetical order, and represent the network splits by concatenating the letters of the blocks from bottom to top, e.g., ``ABCDef" with $L=6$, ``abc" with $L=3$.
	\end{definition}
	
	\begin{definition}[Privatization] \label{def-privatization}
		Given a complete DNN model, we first manually divide it into blocks and obtain the network splits. Then for each $b \in [1, L]$, we take the following four ways to privatize corresponding layers:
		\begin{itemize}
			\item {\bf Private-shared (PS)}. We keep the blocks that lower than $b$ private on local clients and let others participate in the global aggregation.
			\item {\bf Shared-private (SP)}. We keep the $b$th block and upper ones private on local clients and let the lower ones participate in the global aggregation.
			\item {\bf Shared-private-shared (SPS)}. We additionally take a copy of the blocks that lower than $b$ and make them private.
			\item {\bf Shared-shared-private (SSP)}. We additionally take a copy of the $b$th block and upper ones and make them private.
		\end{itemize}
	\end{definition}
	
	We concatenate the letters of the blocks following a bottom-first and shared-first order to name a specific privatization way. For example, ``AaBbCcDE" represents a SPS way that with $L=5$ and $b=4$, ``ABc" represents a SP way that with $L=3$ and $b=3$, and ``ABb" represents a SSP way that with $L=2$ and $b=2$. The illustrations of the above definitions can be found in Fig.~\ref{fig-privatize-teaser}. We only copy the block architecture in SPS and SSP and re-initialize the parameters.
	The SP and PS way can be seen as privatizing the top layers and bottom layers of a network in a single branch style, respectively. Because the server could only access incomplete models from clients, these two ways could not generate a complete global model. Hence, we do not compare aggregation performances with them, and only explore these architectures for personalization. These single branch methods are used in~\cite{FedPer-CoRR2019,LG-FedAvg-CoRR2020}. Different from these two ways, SPS and SSP privatize some layers while keeping a complete model shared across clients. Namely, SPS and SSP are double branch ones\footnote{More branches can also be investigated, and the private branch can also take a different architecture. We do not consider these settings in this paper.}.

	\subsection{Experimental Procedures}
	In this section, we present how to investigate the privatization ways on several benchmarks. Given a DNN based FL scene and its corresponding network, we first divide the network into feature extractor (a.k.a., encoder) and task-specific predictor (a.k.a., classifier). That is, we obtain a type of network splits with $L=2$. Then, according to the defined privatization ways, we can obtain 7 different privatization ways as shown in Fig.~\ref{fig-privatize-teaser}, which are ``AB", ``aB", ``ab", ``Ab", ``AaB", ``AaBb", and ``ABb", respectively. We use ``A'' to denote the shared block of the encoder and ``B'' as the classifier. As aforementioned, lower cases represent private ones. We can observe that ``ab'' works as individually training, ``AB'' works as a single completely shared model used in FedAvg~\cite{FedAvg}, ``aB'' works with a private encoder used in~\cite{LG-FedAvg-CoRR2020}, ``Ab'' works with a private classifier taken in~\cite{FedPer-CoRR2019}, ``AaB'' works as similarly in~\cite{FedFusion,FedMMD},  and ``AaBb'' works similarly as in~\cite{PFL-DA-CoRR2019,FedDML}. Hence, our defined privatization ways can contain existing private-shared FL architectures. For ``AaB'' in local training, we simply take the average of the two encoders' outputs and feed it to the classifier. For ``AaBb'' and ``ABb'', we can obtain two classification probability vectors, and we individually calculate losses of them in training while take their average for prediction as an ensemble in local testing. In FL, we only upload and download the shared blocks as shown in Fig.~\ref{fig-privatize-teaser}.
	
	We will individually apply these privatization ways to the selected benchmarks and compare their FL performances. For each architecture, we also consider taking several groups of hyper-parameters (e.g., the learning rate) and report the best performances.
	Additionally, we will also take a larger $L$ for network splits on several benchmarks and investigate the corresponding performances in a fine-grained privatization way. The results can be found in Sect.~\ref{sect-privatization}.

	\begin{figure}[htbp]
		\centering
		\includegraphics[width=\linewidth]{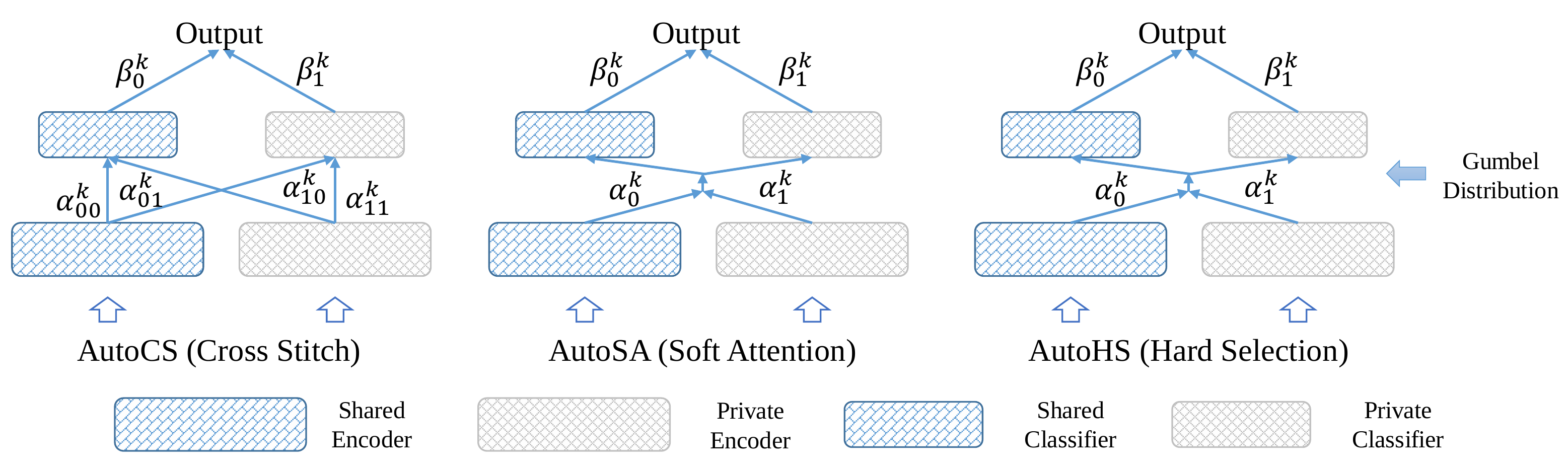}
		\centering
		\caption{{\small Illustrations of the proposed automatical algorithms to learn to aggregate. We only show the architecture and coefficients during local learning procedure on a specific client. The four blocks denote the shared encoder, the shared classifier, the private encoder, and the private classifier, respectively. The $\psi$ in AutoCS, AutoSA, and AutoHS are correspondingly presented in this figure.}}
		\label{fig-auto-learn}
	\end{figure}

	\subsection{Automatically Learn to Aggregate} \label{sect-algos}
	Similar to where to transfer/share in TL/MTL, where to privatize in FL is also a key problem to be solved. In this paper, we propose several automatical algorithms to learn to aggregate in FL. Specifically, we first take a basic architecture as the ``AaBb'', where the global server keeps a global encoder $E_{\ts}$ and a global classifier $C_{\ts}$, and for each $k$, the $k$th client keeps a local encoder $E_{\tp}^k$ and a local classifier $C_{\tp}^k$. The parameters of them are $\te_{E_{\ts}}$, $\te_{C_{\ts}}$, $\te_{E_{\tp}}^k$, and $\te_{C_{\tp}}^k$ correspondingly. To automatically learn a privatization way, some additional parameters should also be globally learned, and we denote them as $\psi$.
	
	In the $t$th round, the server sends the $\te_{E_{\ts},t}$, $\te_{C_{\ts},t}$, and $\psi_{t}$ to the selected clients. Take the $k$th client as an example, it receives these parameters as: $\te_{E_{\ts},t}^k \leftarrow \te_{E_{\ts},t}$, $\te_{C_{\ts},t}^k \leftarrow \te_{C_{\ts},t}$, and $\psi_{t}^k \leftarrow \psi_{t}$. We take the training sample pair $(\bfx_i^k, y_i^k)$ as an example to show the local learning process. The $\bfx_i^k$ will be first processed via the shared and private encoders respectively, i.e., 
	\begin{equation}
		\bfh_{\ts,i}^k \triangleq E_{\ts}^k(\bfx_i^k;\te_{E_{\ts},t}^k), \qquad \bfh_{\tp,i}^k \triangleq E_{\tp}^k(\bfx_i^k;\te_{E_{\tp},t}^k), \label{eq-feat}
	\end{equation}
	where $\bfh_{\ts,i}^k$ and $\bfh_{\tp,i}^k$ denote the shared and private features respectively, and we omit the index of $t$ for simplification. Then we propose three methods to fuse these features, i.e., cross-stitch, soft attention, and hard selection. For cross-stitch, we imitate the learning process in MTL with cross-stitch~\cite{MTL-CrossStitch}, i.e.,
	\begin{equation}
		\hat{\bfh}_{\ts,i}^k \triangleq \hat{\alpha}_{0,0}^k \bfh_{\ts,i}^k + \hat{\alpha}_{0,1}^k \bfh_{\tp,i}^k, \qquad \hat{\bfh}_{\tp,i}^k \triangleq \hat{\alpha}_{1,0}^k \bfh_{\ts,i}^k + \hat{\alpha}_{1,1}^k \bfh_{\tp,i}^k, \label{eq-cross-stitch}
	\end{equation}
	where $\hat{\alpha}_{i,j}^k \in [0,1], i \in \{0,1\}, j \in \{0,1\}$ are coefficients to be learned. In order to obtain a meaningful $\hat{\alpha}^k_{i,j} \in [0, 1]$, we take the softmax to normalize $\alpha_{i,j}^k \in \mathcal{R}$. Specifically, we define $\text{SoftMax}(\cdot, \cdot)$ as a function that takes $\alpha_0$ and $\alpha_1$ as inputs and returns $e^{\alpha_0}/(e^{\alpha_0}+e^{\alpha_1})$ and $e^{\alpha_1}/(e^{\alpha_0}+e^{\alpha_1})$ respectively. Then, we can let $\hat{\alpha}_{0,0}^k,\hat{\alpha}_{0,1}^k=\text{SoftMax}(\alpha_{0,0}^k / \lambda, \alpha_{0,1}^k / \lambda)$, and $\hat{\alpha}_{1,0}^k,\hat{\alpha}_{1,1}^k=\text{SoftMax}(\alpha_{1,0}^k / \lambda, \alpha_{1,1}^k / \lambda)$. $\lambda$ is the temperature. In our experiments, taking it as $2.0$ could lead to slightly better results.
	
	For soft attention, we take a simple weighted average of the features as follows:
	\begin{equation}
		\hat{\bfh}_{\ts,i}^k \triangleq \hat{\alpha}_{0}^k \bfh_{\ts,i}^k + \hat{\alpha}_{1}^k \bfh_{\tp,i}^k, \qquad \hat{\bfh}_{\tp,i}^k \triangleq \hat{\alpha}_{0}^k \bfh_{\ts,i}^k + \hat{\alpha}_{1}^k \bfh_{\tp,i}^k, \label{eq-soft-attention}
	\end{equation}
	where we can find that this differs from cross-stitch only in that it takes the same weights to generate $\hat{\bfh}_{\ts,i}^k$ and $\hat{\bfh}_{\tp,i}^k$. Similarly, we take $\hat{\alpha}_{0}^k,\hat{\alpha}_{1}^k=\text{SoftMax}(\alpha_{0}^k / \lambda, \alpha_{1}^k / \lambda)$. We propose this approach to investigate whether the symmetric property of the combination weights matters.
	
	For hard selection, we directly select one group of features via the Gumbel-softmax~\cite{Gumbel-Softmax,DMIS}, i.e.,
	\begin{equation}
		\hat{\bfh}_{\ts,i}^k \triangleq \hat{\alpha}_{0}^k \bfh_{\ts,i}^k + \hat{\alpha}_{1}^k \bfh_{\tp,i}^k, \qquad \hat{\bfh}_{\tp,i}^k \triangleq \hat{\alpha}_{0}^k \bfh_{\ts,i}^k + \hat{\alpha}_{1}^k \bfh_{\tp,i}^k, \label{eq-hard-selection}
	\end{equation}
	where $\hat{\alpha}_{0}^k$ and $\hat{\alpha}_{1}^k$ can be generated via the following steps. First, sample values from the Gumbel distribution, i.e., $g_0, g_1 \sim \text{Gumbel}(0,1)$, where $\text{Gumbel}(0,1)$ can be seen as first sampling $u_0, u_1 \sim \text{Uniform}(0,1)$ and then taking the transformation $g = -\log(-\log(u))$. Second, take the reparametrized softmax as $\hat{\alpha}_0^k, \hat{\alpha}_1^k = \text{SoftMax}((\alpha_0^k + g_0)/\lambda, (\alpha_1^k + g_1)/\lambda)$. We denote the process as $\hat{\alpha}_{0}^k,\hat{\alpha}_{1}^k=\text{GumbelSoftMax}(\alpha_{0}^k, \alpha_{1}^k)$. Using Gumbel softmax can lead to more spiked weights or even discrete selection codes~\cite{Gumbel-Softmax}, and hence we expect it could lead to purer aggregation and more understandable results.
	
	With either of the three ways, we can obtain $\hat{\bfh}_{\ts,i}^k$ and $\hat{\bfh}_{\tp,i}^k$. Then we make predictions via:
	\begin{equation}
		\bfo_{\ts,i}^k \triangleq C_{\ts}^k(\hat{\bfh}_{\ts,i}^k;\te_{C_{\ts},t}^k), \qquad \bfo_{\tp,i}^k \triangleq C_{\tp}^k(\hat{\bfh}_{\tp,i}^k;\te_{C_{\tp},t}^k), \label{eq-classify}
	\end{equation}
	where $\bfo_{\ts,i}^k$ and $\bfo_{\tp,i}^k$ are the unnormalized outputs of shared classifier and private classifier correspondingly. Before using softmax and calculating cross-entropy loss, we similarly take a symmetric fusion as follows:
	\begin{equation}
		\hat{\bfo}_{i}^k \triangleq \hat{\beta}_{0}^k \bfo_{\ts,i}^k + \hat{\beta}_{1}^k \bfo_{\tp,i}^k, \label{eq-fuse-prediction}
	\end{equation}
	where $\hat{\bfo}_{i}^k$ is the final weighted output, and it will be applied with softmax to calculate the cross-entopy loss. For cross-stitch and soft attention, $\hat{\beta}_{0}^k, \hat{\beta}_{1}^k = \text{SoftMax}(\beta_{0}^k / \lambda, \beta_{1}^k / \lambda)$, while for hard selection, $\hat{\beta}_{0}^k, \hat{\beta}_{1}^k = \text{GumbelSoftMax}(\beta_{0}^k, \beta_{1}^k)$.
	
	In total, for cross-stitch, we have $\psi=\{\alpha_{0,0},\alpha_{0,1},\alpha_{1,0},\alpha_{1,1},\beta_{0},\beta_{1}\}$; for soft attention or hard selection, we have $\psi=\{\alpha_{0},\alpha_{1},\beta_{0},\beta_{1}\}$. These parameters will be end-to-end updated in local training and will also be aggregated during the global procedure. The three algorithms are named as AutoCS, AutoSA, and AutoHS respectively, whose pseudo codes can be found in Algo.~\ref{algo}. The $\psi$ to be learned and the architectures are shown in Fig.~\ref{fig-auto-learn}.

	\begin{table}[htbp]
		\centering
		\caption{{\small Statistical information of the investigated benchmarks: the number of clients $K$, the total number of classes $C$, the number of training samples on each client on average $\overline{\Nktr}$, the number of test samples on each client on average $\overline{\Nkte}$, and the number of global test samples $M$.}} \label{tab:scene-detail}
		\vspace{-0.3cm}
		{
			\begin{tabular}{l|c|c|c|c|c||l|c|c|c|c|c}
				\hline \hline
				& $K$ & $C$ & $\overline{\Nktr}$ & $\overline{\Nkte}$ & $M$ & & $K$ & $C$ & $\overline{\Nktr}$ & $\overline{\Nkte}$ & $M$ \\
				\hline \hline
				FeCifar10 & 100 & 10 & 400 & 100 & 10k & FeCifar100 & 100 & 100 & 400 & 100 & 10k \\
				\hline
				Shakespeare & 1129 & 81 & 2994 & 749 & 845k & FeMnist & 3550 & 62 & 181 & 45 & 161k \\
				\hline \hline
			\end{tabular}
		}
		\vspace{-0.3cm}
	\end{table} 
	
	\begin{figure}[htbp]
		\centering
		
		\begin{minipage}{0.45\linewidth}
			\centering
			\includegraphics[width=\linewidth]{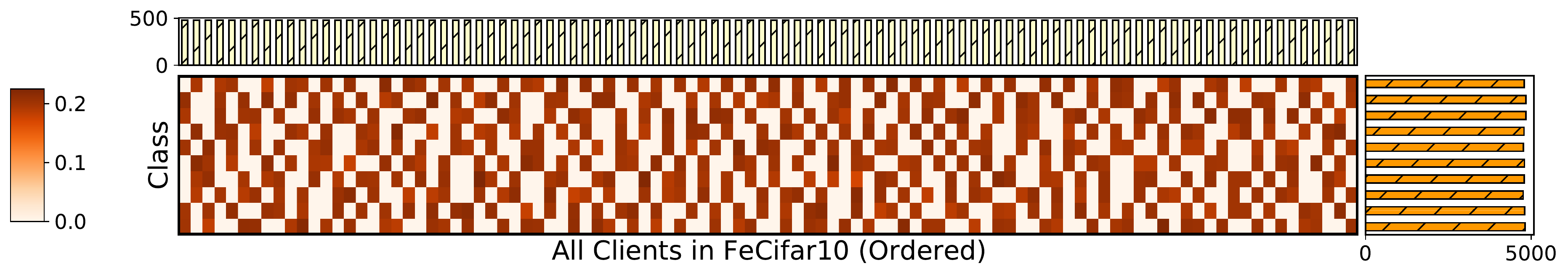}
			\centerline{(a) FeCifar10}
		\end{minipage}
		\begin{minipage}{0.45\linewidth}
			\centering
			\includegraphics[width=\linewidth]{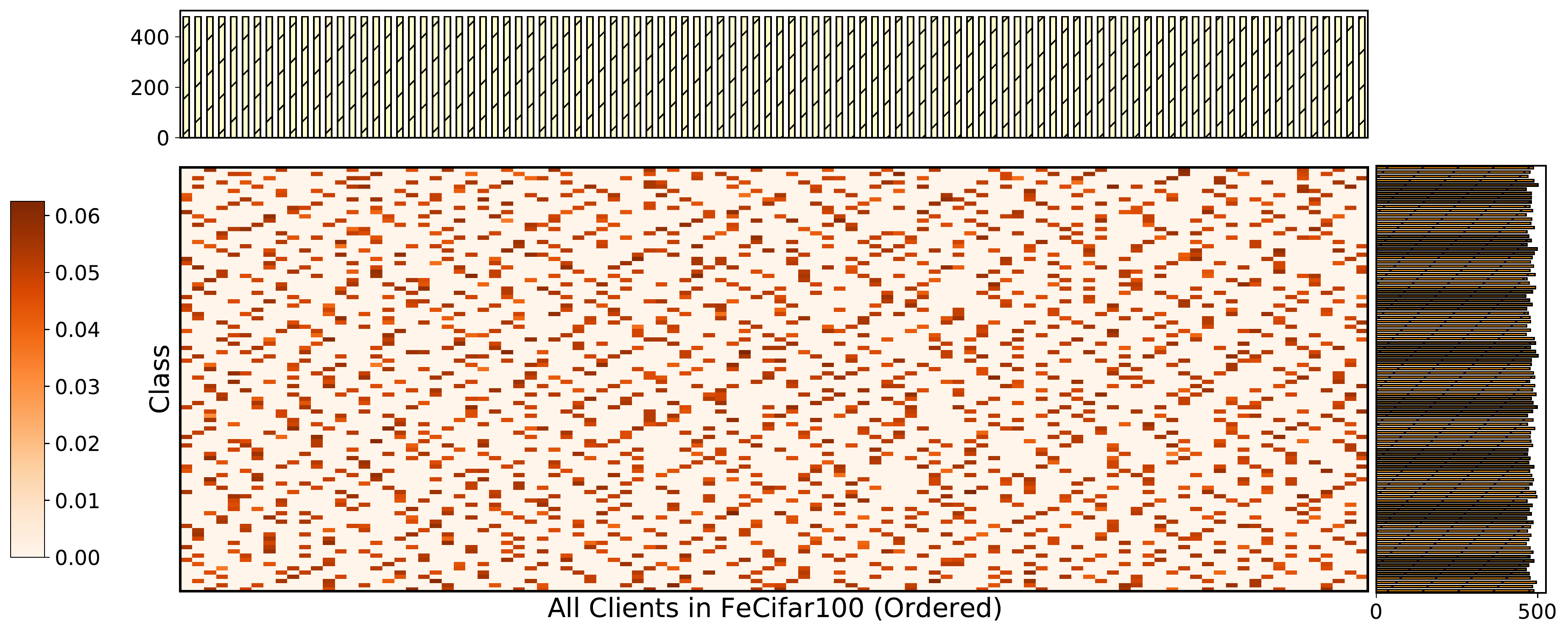}
			\centerline{(b) FeCifar100}
		\end{minipage}
		\begin{minipage}{0.45\linewidth}
			\centering
			\includegraphics[width=\linewidth]{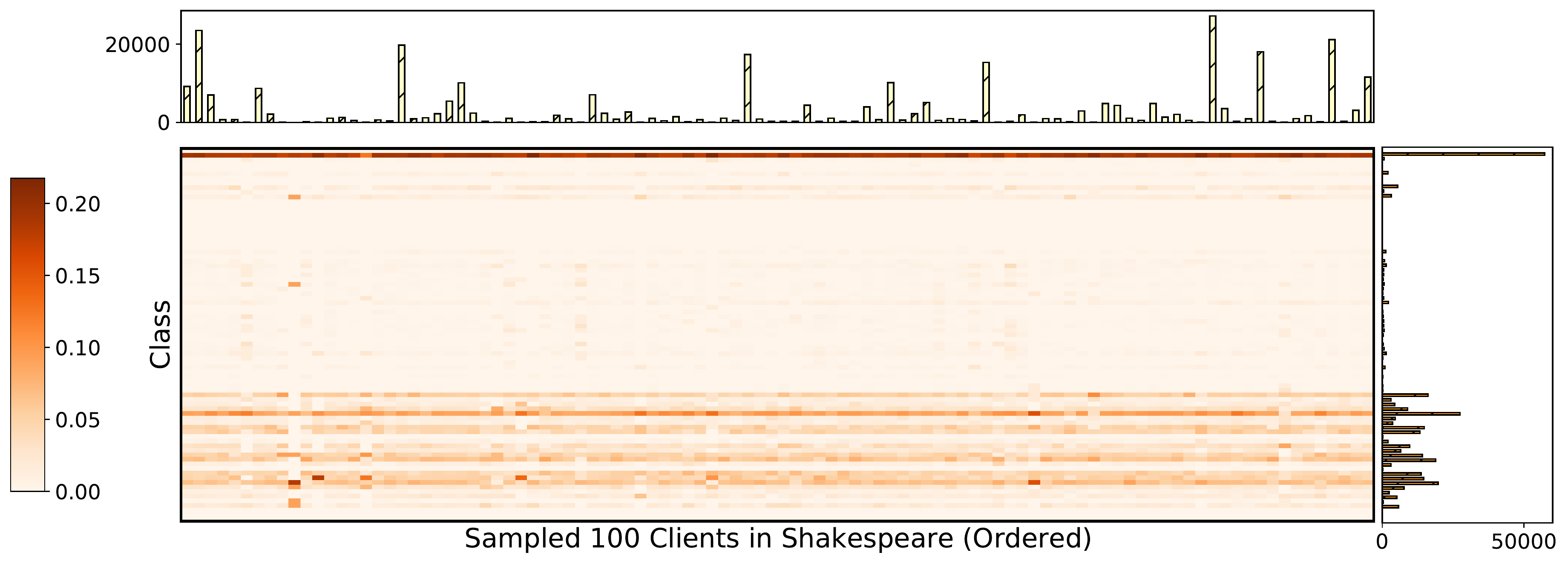}
			\centerline{(c) Shakespeare}
		\end{minipage}
		\begin{minipage}{0.45\linewidth}
			\centering
			\includegraphics[width=\linewidth]{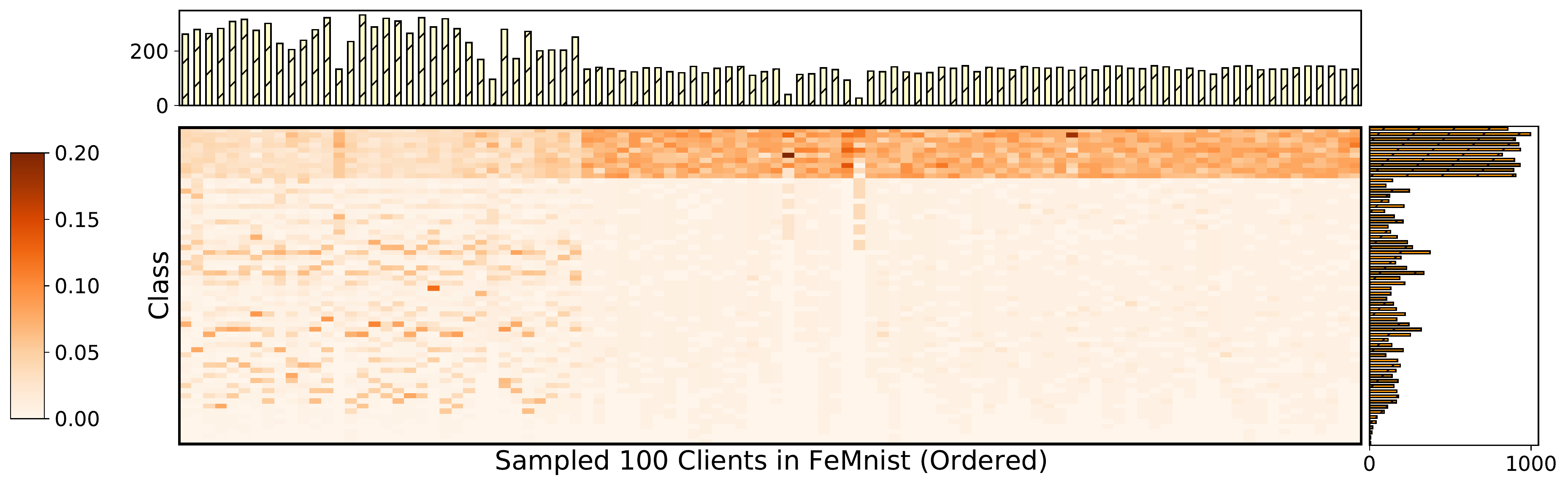}
			\centerline{(d) FeMnist}
		\end{minipage}
		
		\centering
		\caption{{\small Distributions of the investigated FL benchmarks. Every figure shows one benchmark. In each figure, the middle heatmap shows the label distributions of 100 clients; the top histogram plots the numbers of samples in each client; the right histogram shows the number of samples in each class.}}
		\label{fig-info}
	\end{figure}
	
	\section{Experimental Studies}
	In this section, we will detail our experimental studies. We first introduce the utilized four benchmarks and their statistics. Then, we will present the utilized networks for these benchmarks. Finally, we will report the detailed experimental results.
	
	\subsection{Benchmarks}
	We select several FL benchmarks. In order to represent both covariate shift and label shift non-iid scenes~\cite{Fed-Advances}, we use FeCifar10, FeCifar100, Shakespeare, and FeMnist. The FeCifar10 and FeCifar100 are constructed by splitting cifar10 and cifar100 dataset~\cite{cifar} according to labels. Specifically, for FeCifar10, we split the training samples of each class in cifar10 into 50 shards and obtain $10 \times 50 = 500$ shards in all, then we class-wisely allocate 5 shards to each client and each client could access 5 different classes. For FeCifar100, we split the training samples of each class in cifar100 into 20 shards and obtain $100 \times 20 = 2000$ shards in all. We class-wisely allocate 20 shards to each client and each client could access 20 different classes. That is, we have 100 clients in both FeCifar10 and FeCifar100. This way of decentralization is mainly based on the label distribution, and we advocate these two scenes are more inclined to be label distribution shift non-iid scenes. Such partitions can also be found in previous works~\cite{Fed-NonIID-Data,FedMD}. Shakespeare and FeMnist are benchmarks recommended by LEAF~\cite{LEAF}. Shakespeare is a dataset built from the Complete Works of William Shakespeare, which is originally used in FedAvg~\cite{FedAvg}. It is constructed by viewing each speaking role in each play as a different device, and the target is to predict the next character based on the previous characters. FeMnist is a task to classify the mixture of digits and characters, where data from each writer is considered as a client. Different speakers or writers may behave distinctly with various semantics and styles. Hence, we view these two scenes as covariate shift non-iid ones, where the heterogeneity is mainly caused by different play roles or users.
	
	For FeCifar10 and FeCifar100, we split the local data into 80\% and 20\% as local train set and local test set respectively, and use the original test set in cifar10 and cifar100 as the global test set. For Shakespeare and FeMnist, we split local data into 80\% and 20\% as local train set and local test set respectively, while we take the combination of all local test sets as the global test set. We reshape the images in FeMnist to $28\times 28$. We list the details of the benchmarks from several aspects in Tab.~\ref{tab:scene-detail}: the number of clients $K$, the total number of classes $C$, the number of training samples on each client on average $\overline{\Nktr}$, the number of test samples on each client on average $\overline{\Nkte}$, and the number of global test samples $M$.
	
	For a deeper understanding of these scenes, we plot the distributions of some clients in Fig.~\ref{fig-info}. FeCifar10 and FeCifar100 are artificially constructed label shift non-iid scenes, where each client could only access a subset of classes, and the label distributions among clients differ a lot. Although some classes are missing for a specific client, the observed classes are uniformly distributed. Additionally, the samples are uniformly distributed across clients and classes. Shakespeare and FeMnist also experience a certain level of label shift, while the label distributions among clients do not diverge too much and are almost consistent with the global label distribution. Compared with FeCifar10 and FeCifar100, Shakespeare and FeMnist have another two challenges, i.e., the imbalance of the sample numbers across clients and classes. For example, some clients in Shakespeare could have more than 20,000 training samples, and the first 10 classes (i.e., digits) in FeMnist dominate most of the samples. However, we do not additionally solve these problems in this paper. We only suggest that some corresponding solutions (e.g., FL with imbalance learning techniques~\cite{FL-Imbalance}) can be adopted to improve performance.
	
	\begin{figure}[htbp]
		\centering
		\includegraphics[width=\linewidth]{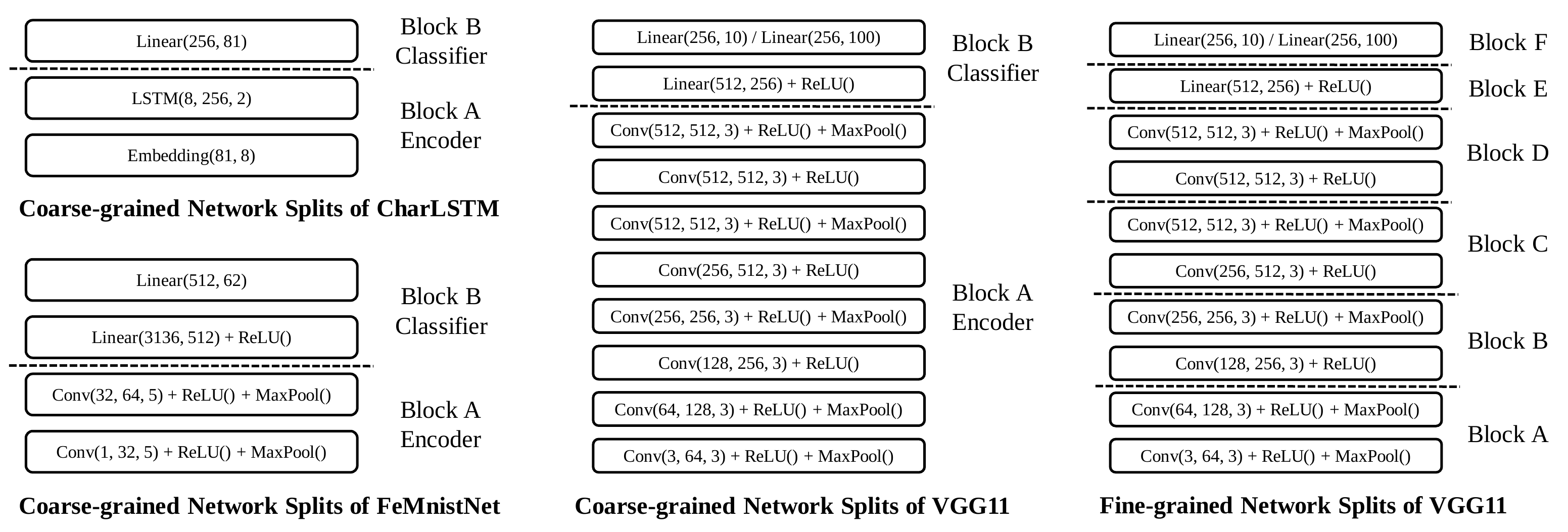}
		\centering
		\caption{{\small Illustrations of the utilized networks and corresponding network splits. The left two columns show the coarse-grained network splits of CharLSTM (for Shakespeare), FeMnistNet (for FeMnist), and VGG11 (for FeCifar10 and FeCifar100). The right shows the fine-grained network splits of VGG11. The three numbers in ``Conv()'' denote the input channel, output channel, and kernel size, respectively; the ``MaxPool()'' has a default kernel size as 2; the two numbers in ``Linear()'' denote the input and output size; the ``LSTM(8, 256, 2)'' denotes the embedding size is 8, the hidden size is 256, and the number of layers is 2.}}
		\label{fig-networks}
	\end{figure}
	
	\subsection{Utilized Networks and Network Splits}
	We investigate both coarse-grained and fine-grained privatization ways. For coarse-grained investigation, we split the utilized network into $L=2$ blocks, i.e., the encoder and classifier. For FeCifar10 and FeCifar100, we utilize the VGG11 without batch normalization in PyTorch\footnote{\url{https://pytorch.org/vision/master/_modules/torchvision/models/vgg.html}}. For Shakespeare, we take the same network used in FedAvg~\cite{FedAvg}, which contains a character embedding layer, a two-layer LSTM, and a fully-connected layer. For FeMnist, we utilize the FeMnistNet as used in LEAF~\cite{LEAF}, which contains two convolution layers and two fully-connected layers. Owing to the memory limitation, we modify the first fully-connected layer's output dimension from 2048 to 512. For all networks, we view the fully-connected layers as the classifier and others as the encoder as shown in Fig.~\ref{fig-networks}. For fine-grained network splits, we only investigate the performances on FeCifar10 and FeCifar100, splitting the VGG11 into $L=6$ blocks as shown in Fig.~\ref{fig-networks}.
	
	\begin{figure}[htbp]
		\centering
		
		\begin{minipage}{0.45\linewidth}
			\centering
			\includegraphics[width=\linewidth]{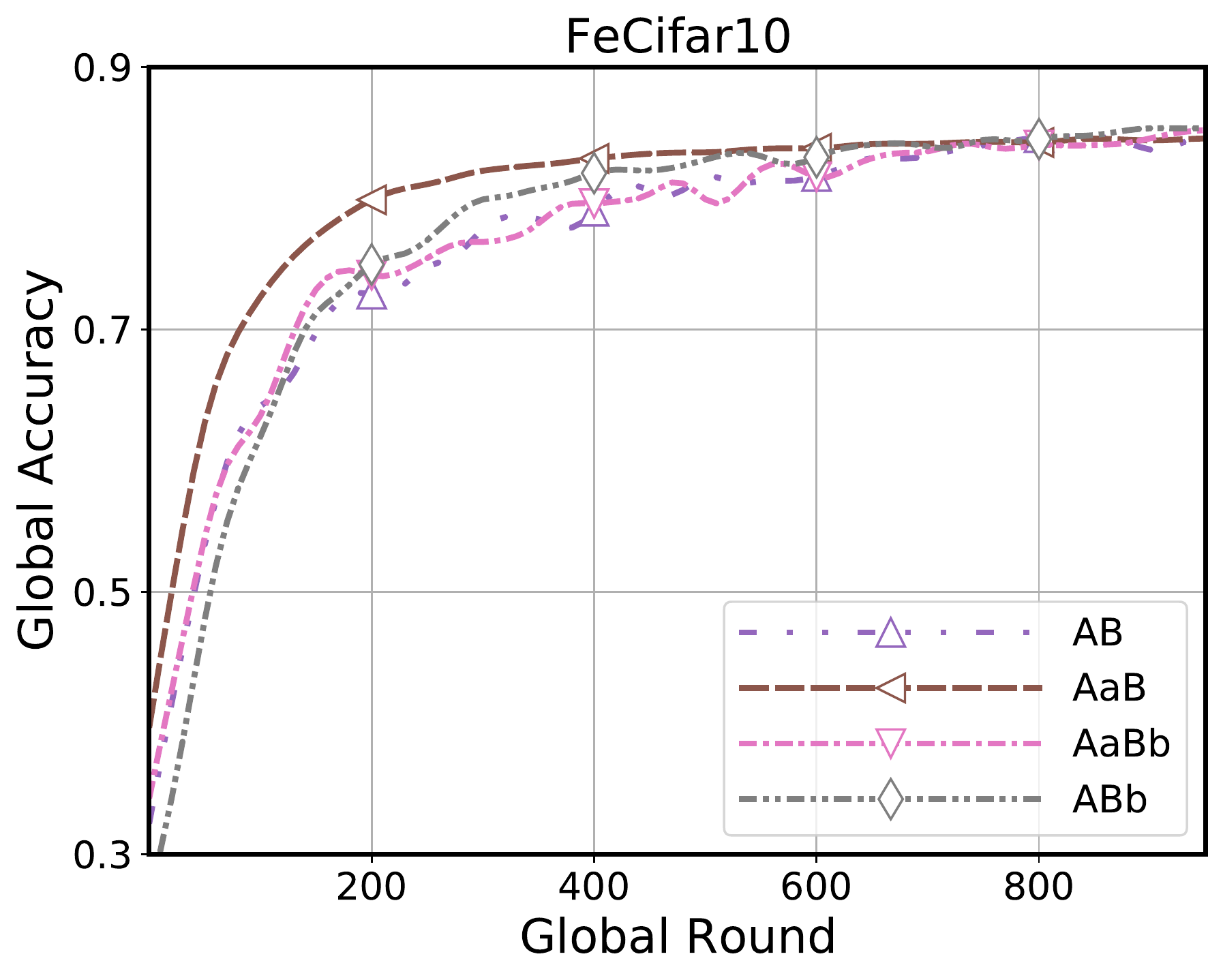}
			\centerline{(a) FeCifar10}
		\end{minipage}
		\quad
		\begin{minipage}{0.45\linewidth}
			\centering
			\includegraphics[width=\linewidth]{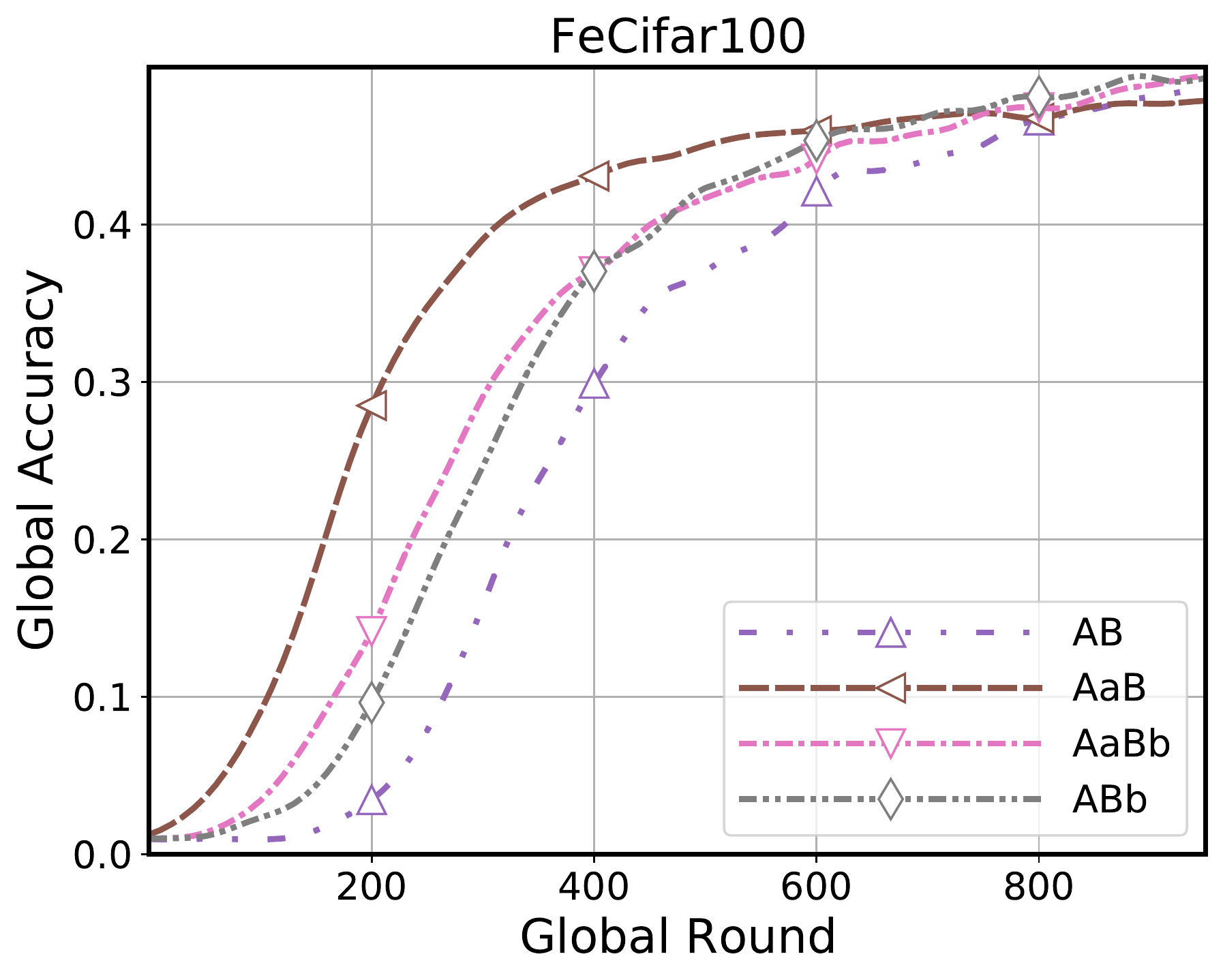}
			\centerline{(b) FeCifar100}
		\end{minipage}
		\begin{minipage}{0.45\linewidth}
			\centering
			\includegraphics[width=\linewidth]{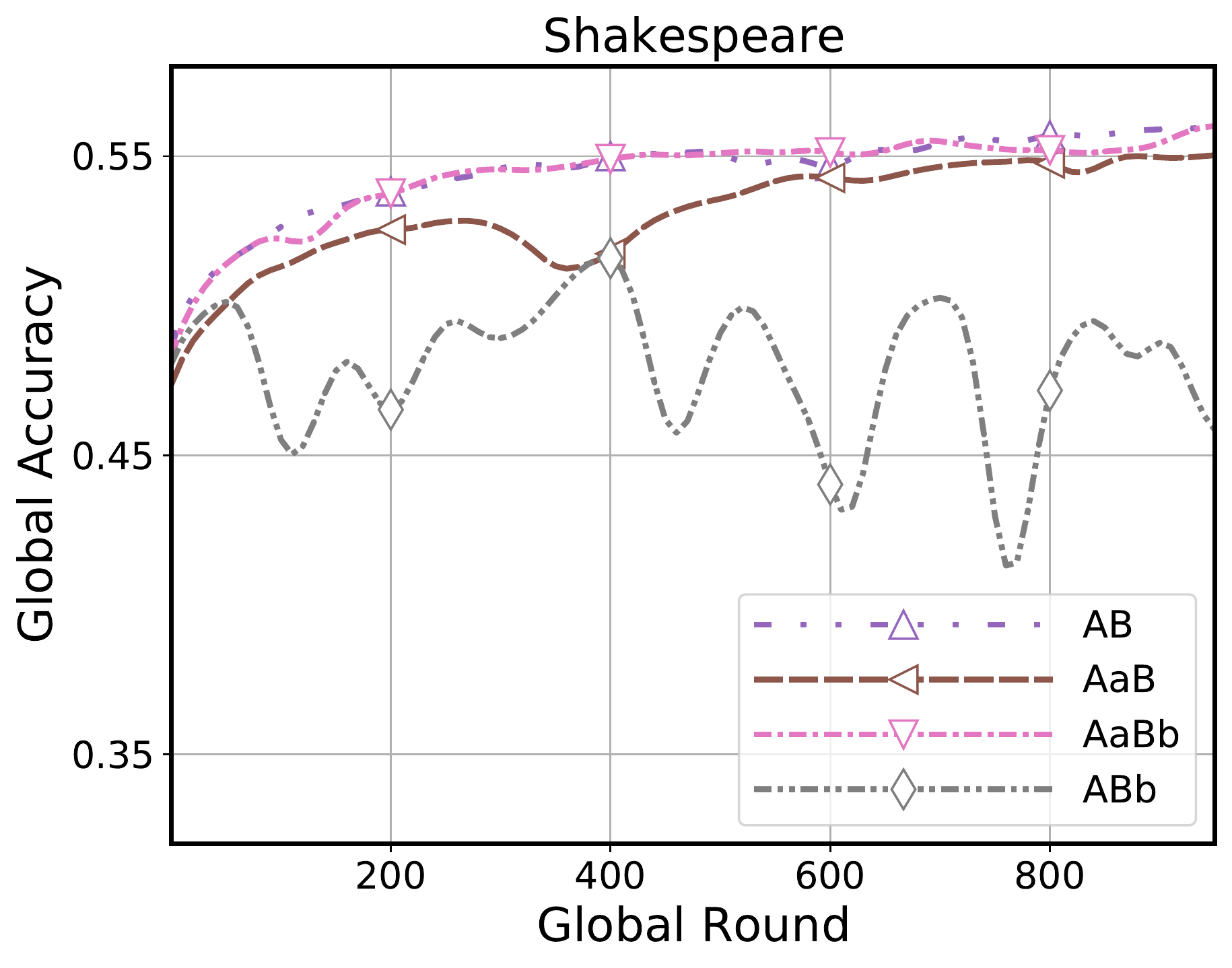}
			\centerline{(c) Shakespeare}
		\end{minipage}
		\quad
		\begin{minipage}{0.45\linewidth}
			\centering
			\includegraphics[width=\linewidth]{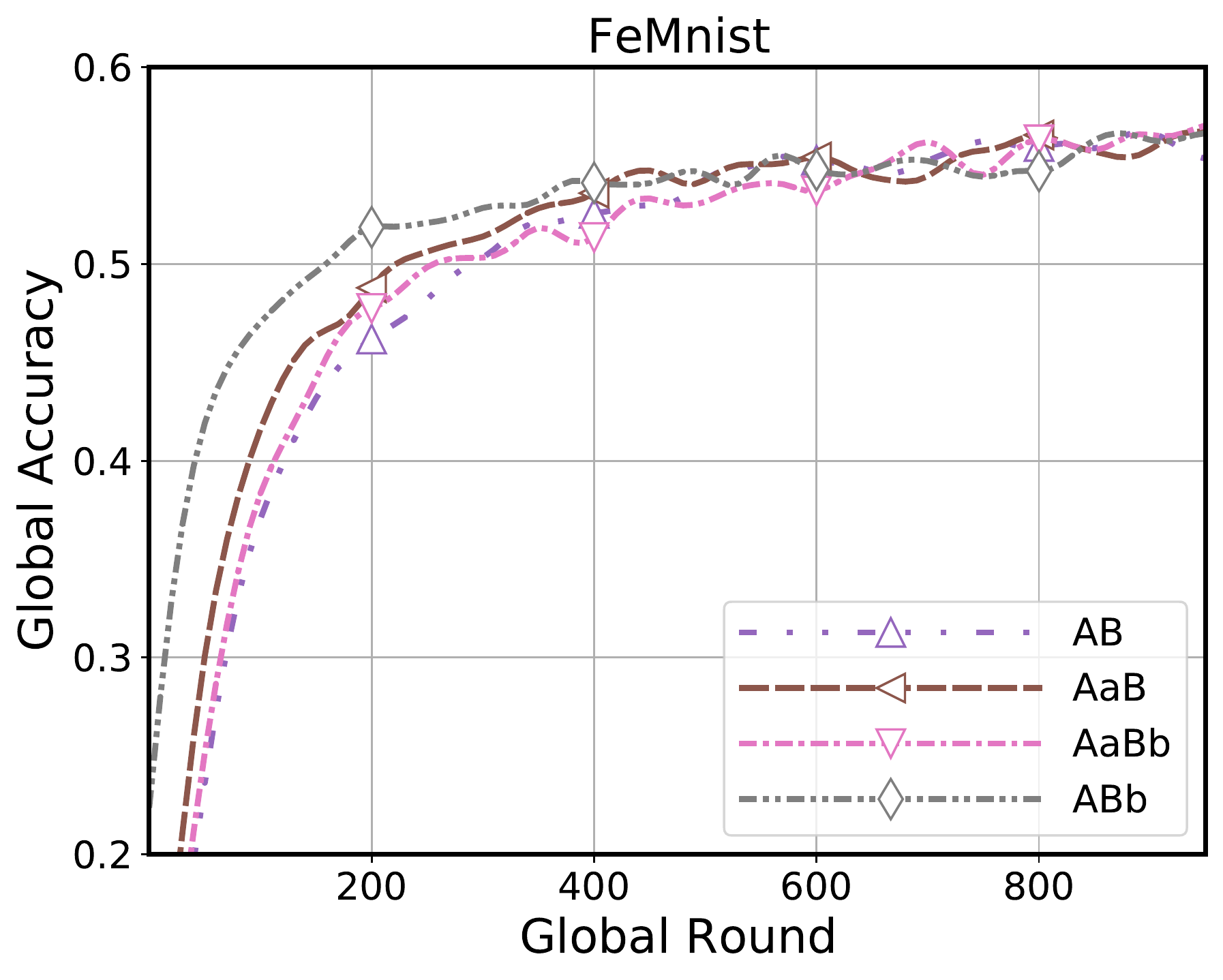}
			\centerline{(d) FeMnist}
		\end{minipage}
		
		\centering
		\caption{{\small Aggregation performances of different privatization ways under the four non-iid benchmarks. Every figure shows one benchmark.}}
		\label{fig-coarse-agg}
	\end{figure}
	
	\begin{figure}[htbp]
		\centering
		
		\begin{minipage}{0.45\linewidth}
			\centering
			\includegraphics[width=\linewidth]{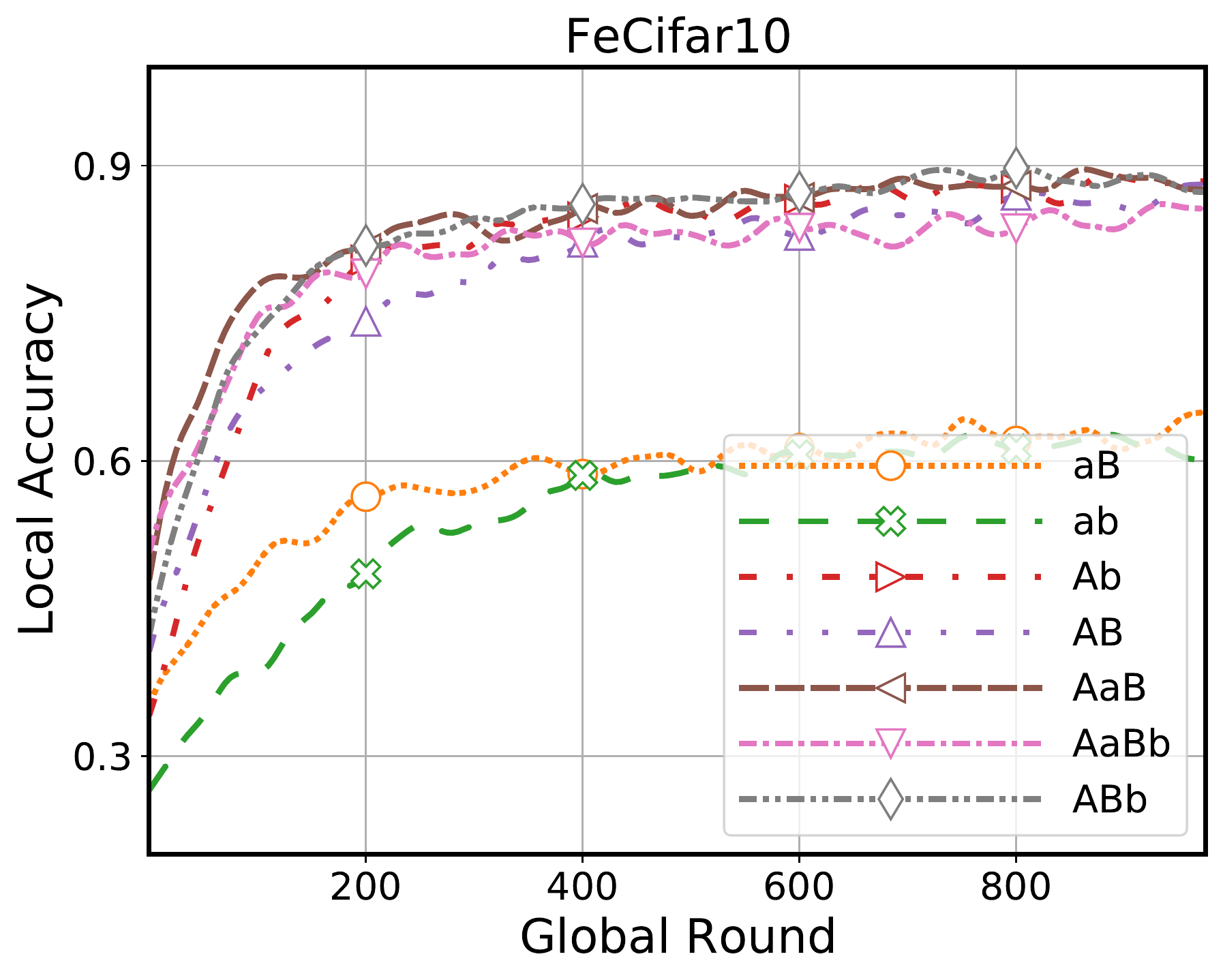}
			\centerline{(a) FeCifar10}
		\end{minipage}
		\quad
		\begin{minipage}{0.45\linewidth}
			\centering
			\includegraphics[width=\linewidth]{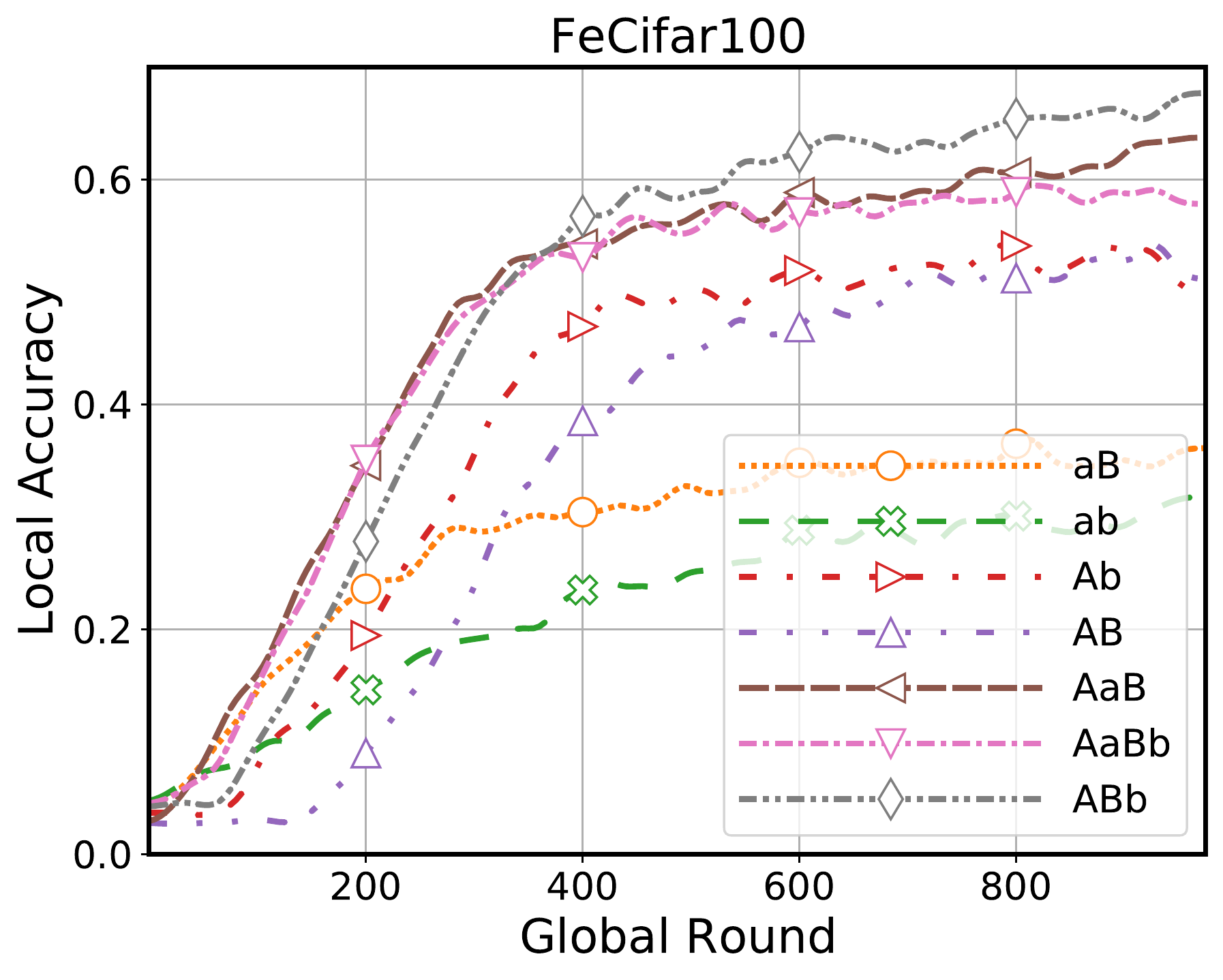}
			\centerline{(b) FeCifar100}
		\end{minipage}
		\begin{minipage}{0.45\linewidth}
			\centering
			\includegraphics[width=\linewidth]{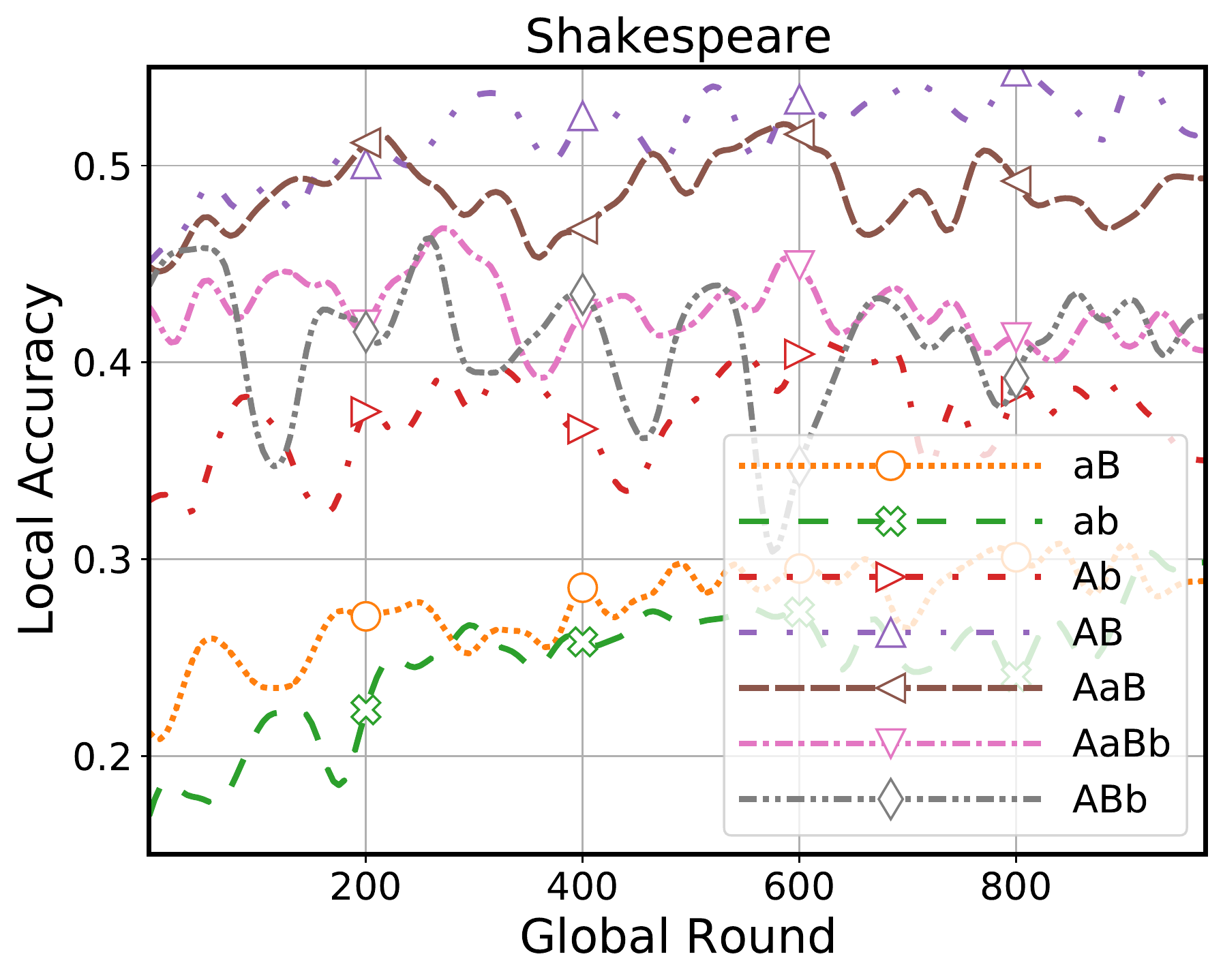}
			\centerline{(c) Shakespeare}
		\end{minipage}
		\quad
		\begin{minipage}{0.45\linewidth}
			\centering
			\includegraphics[width=\linewidth]{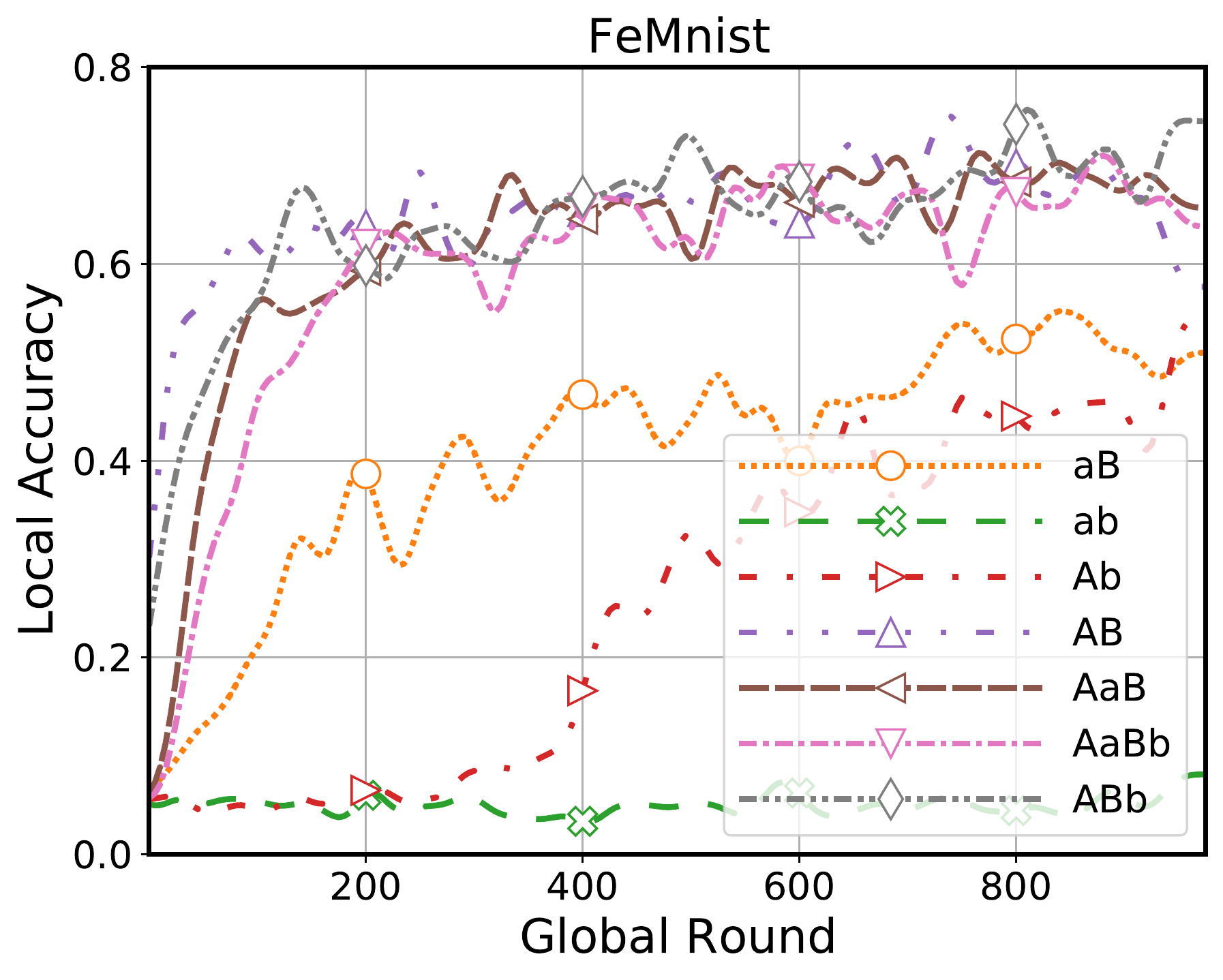}
			\centerline{(d) FeMnist}
		\end{minipage}
		
		\centering
		\caption{{\small Personalization performances of different privatization ways under the four non-iid benchmarks. Every figure shows one benchmark.}}
		\label{fig-coarse-per}
	\end{figure}

	\subsection{Hyper-parameters and Settings}
	We denote the learning rate as $\eta$, the batch size as $B$,  the client selection ratio as $Q$, the maximum number of global round as $T$, and the local training epochs as $E$. For FeCifar10 and FeCifar100, we vary $\eta$ in $\{0.01, 0.03, 0.05\}$, take $B=64$, $Q=0.1$, $T=1000$, and $E=2$; for Shakespeare, we vary $\eta$ in $\{0.75, 1.0, 1.47\}$, take $B=50$, $Q=0.01$, $T=1000$, and $E=1$; for FeMnist, we vary $\eta$ in $\{0.001, 0.002, 0.004\}$, take $B=10$, $Q=0.001$, $T=1000$, and $E=2$. For all of them, we use SGD with momentum 0.9 as the optimizer. For shakespeare, some clients have amounts of training samples, and hence we limit the maximum of local steps (i.e., the number of local batches) to be $250$. In all experiments, we record both the aggregation and personalization results as calculated in Def.~\ref{def-aggregation} and Def.~\ref{def-personalization} every 10 rounds. For various hyper-parameters, we report the best one according to the average of the last 5 recorded performances. Taking various hyper-parameters can omit the randomness in training to some extent, and the obtained results are more convincing.
	
	\begin{figure}[htbp]
		\centering
		
		\begin{minipage}{0.45\linewidth}
			\centering
			\includegraphics[width=\linewidth]{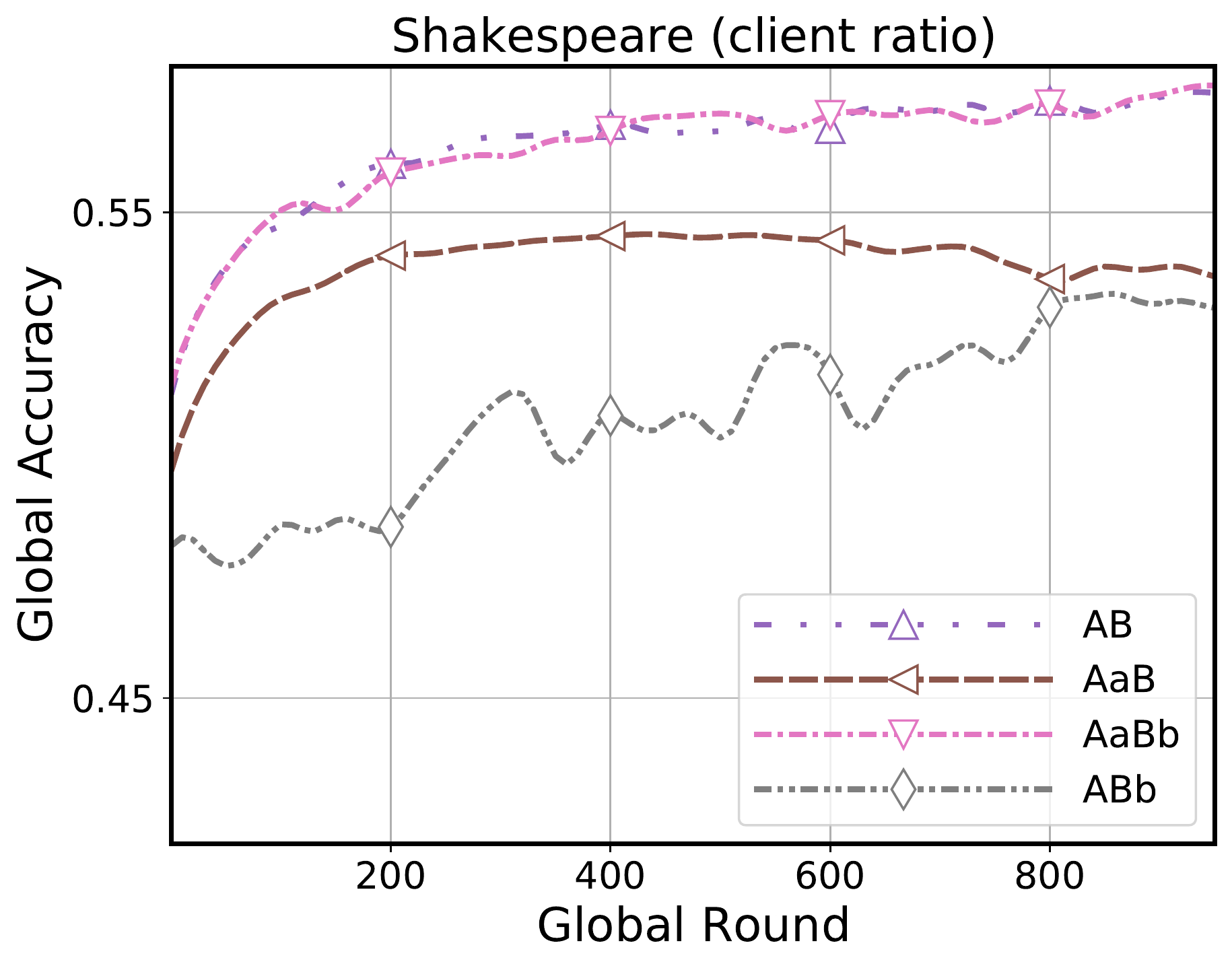}
			\centerline{(a) Shakespeare, Aggregation ($Q=0.1$)}
		\end{minipage}
		\quad
		\begin{minipage}{0.45\linewidth}
			\centering
			\includegraphics[width=\linewidth]{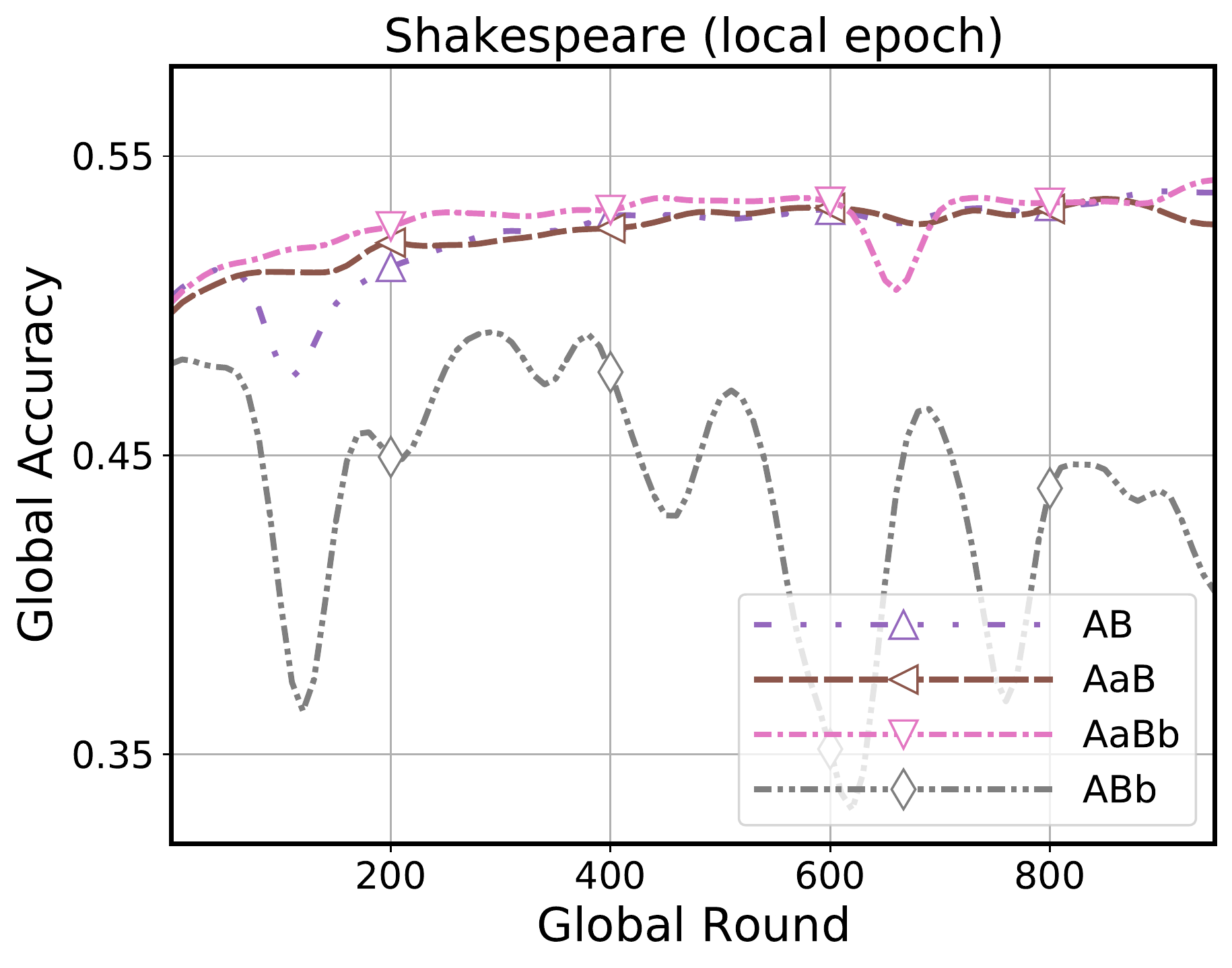}
			\centerline{(b) Shakespeare, Aggregation ($E=5$)}
		\end{minipage}
		
		\begin{minipage}{0.45\linewidth}
			\centering
			\includegraphics[width=\linewidth]{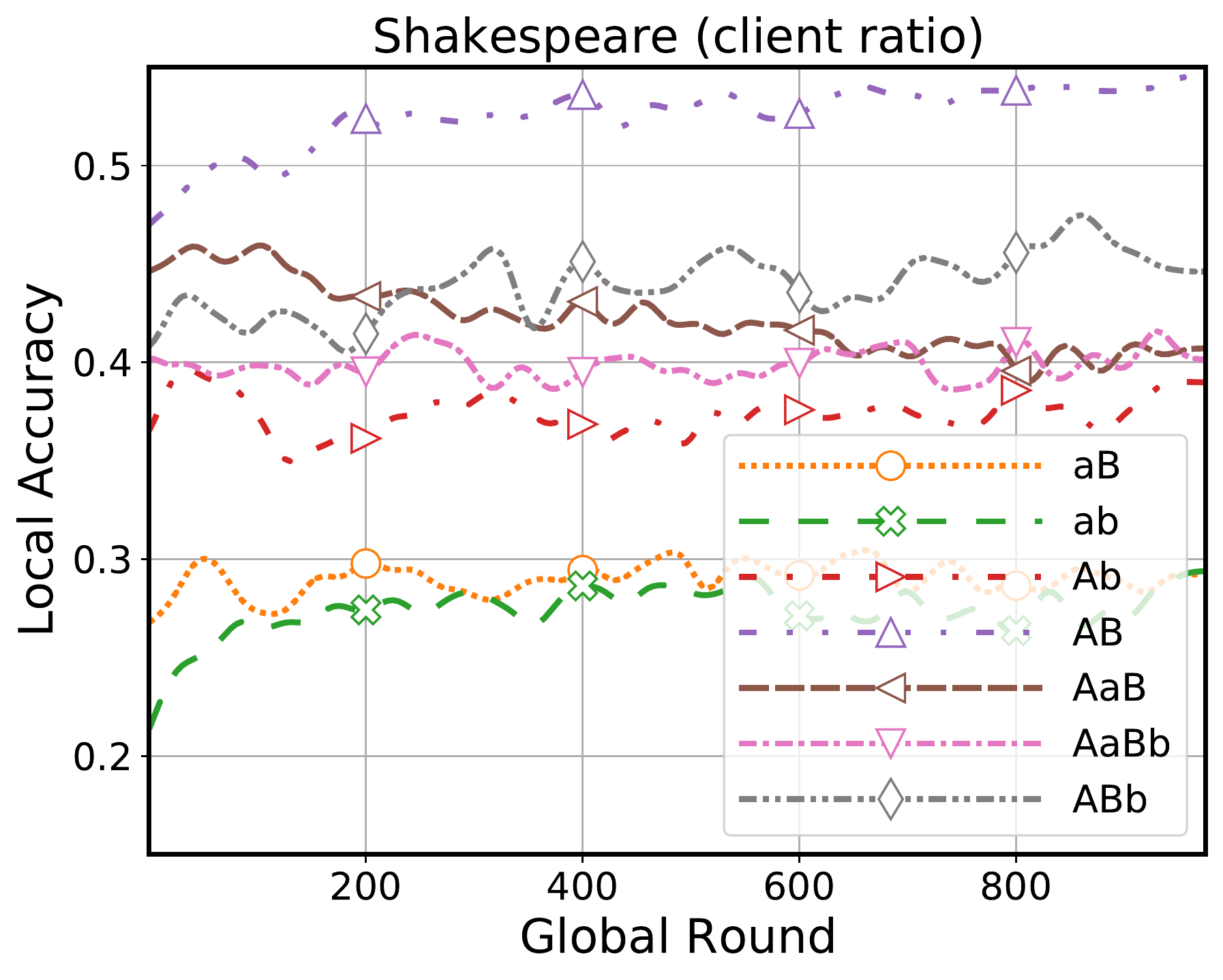}
			\centerline{(c) Shakespeare, Personalization ($Q=0.1$)}
		\end{minipage}
		\quad
		\begin{minipage}{0.45\linewidth}
			\centering
			\includegraphics[width=\linewidth]{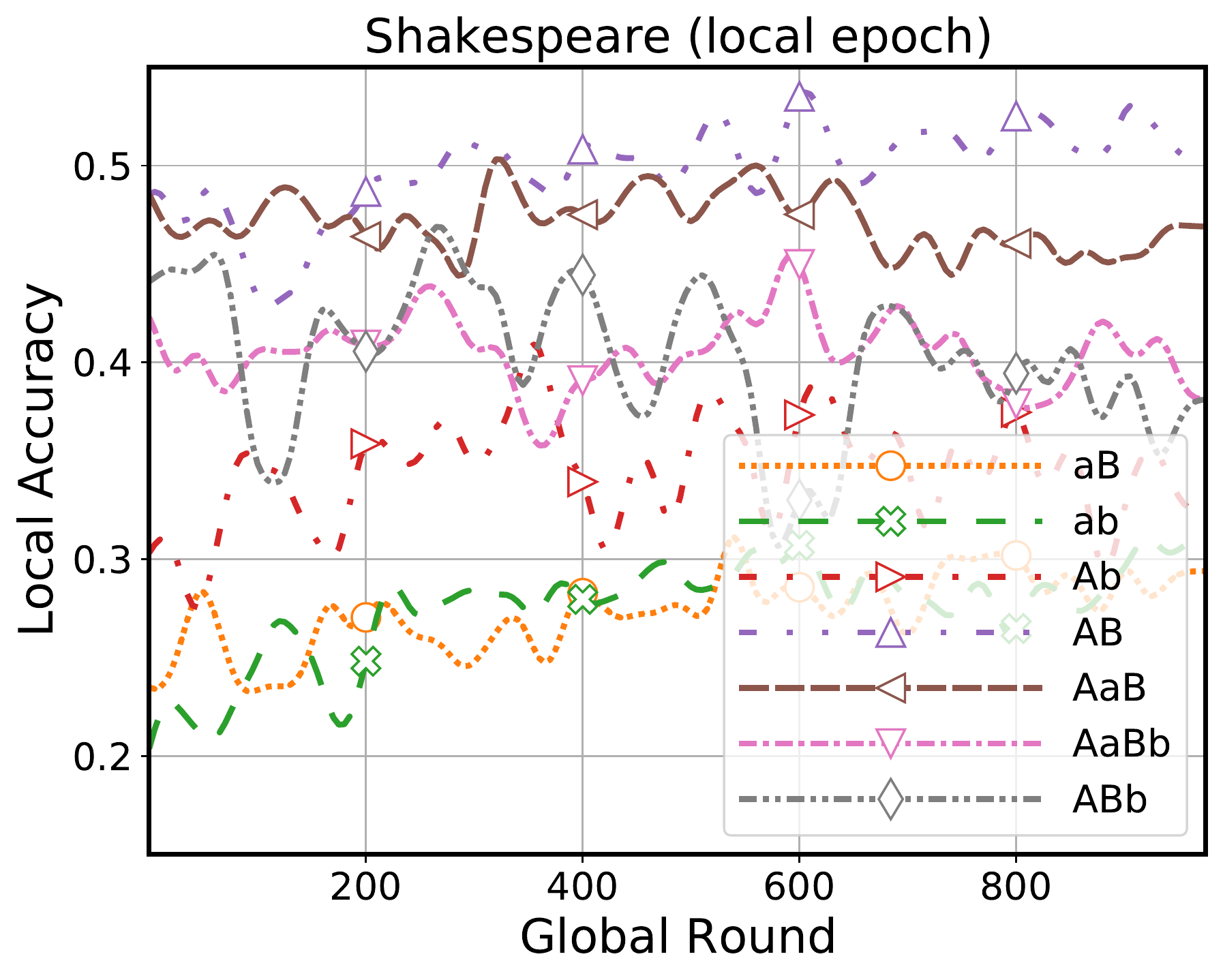}
			\centerline{(d) Shakespeare, Personalization ($E=5$)}
		\end{minipage}
		
		\centering
		\caption{{\small The aggregation and personalization performances of Shakespeare under various hyper-parameter settings. The top row shows the aggregation performances, while the bottom shows the personalization performances. In each row, the two figures show the case when the client selection ratio is 0.1, and the number of local epochs is 5, respectively.}}
		\label{fig-shake-more}
	\end{figure}

	\subsection{Results of Different Privatization Ways} \label{sect-privatization}
	\subsubsection{Coarse-grained Privatization} \label{sect-coarse-privatization}
	We first report the results of the coarse-grained privatization ways. We split corresponding networks into two blocks, i.e., encoder and classifier, as shown in Fig.~\ref{fig-privatize-teaser}. We report the aggregation and personalization results in Fig.~\ref{fig-coarse-agg} and Fig.~\ref{fig-coarse-per}, respectively. We denote the aggregation performance as ``Global Accuracy'' and the personalization performance as ``Local Accuracy'' in these figures. In Fig.~\ref{fig-coarse-agg}, we can observe that ``AaB'' can lead to much better aggregation performances on FeCifar10 and FeCifar100, which contradicts the two assumptions in Sect.~\ref{sect-issues}. FeCifar10 and FeCifar100 are constructed via splitting labels, which may perform better with ``ABb'' if the two assumptions stand. However, we find that ``ABb'' performs similarly to ``AaBb'', a little better than ``AB'', while worse than ``AaB''. This phenomenon is not caused by the number of parameters because their parameters are the same on the server. Then, we consider the personalization performances as in Fig.~\ref{fig-coarse-per}, the ``ABb'' can lead to comparable or a little better results than ``AaB'' and ``AaBb'' on FeCifar10 and FeCifar100. The individually training, i.e., ``ab'', obtains the worst personalization accuracies, which is rational because of the limit of training samples on local clients. Privatizing the encoder in a single branch style, i.e., ``aB'', works better than ``ab'' while worse than ``AB''. However, privatizing the classifier, i.e., ``Ab'', can obtain slightly better performances than ``AB''. To be brief, on FeCifar10 and FeCifar100, privatizing an encoder with double branches, i.e., ``AaB'', can lead to better aggregation and personalization performances. {\em Namely, for FL with label shift non-iid scenes, privatizing a classifier or a complete model may be less effective than privatizing an encoder, which contradicts some natural conjectures.}
	
	Then, we take a look at Shakespeare and FeMnist. {\em On Shakespeare, all privatization ways seem useless in both aggregation and personalization performances.} This is so exciting that we doubt whether Shakespeare is a real non-iid scene or not, and we will verify our assumption later. On FeMnist, ``ABb" can lead to faster convergence for aggregation, and the following is ``AaB'', while they perform comparably for personalization. This phenomenon contradicts the first assumption in Sect.~\ref{sect-issues}, where the covariate shift non-iid scene does not necessarily require privatizing an encoder.
	The above observations can be concluded as follows:
	\begin{itemize}
		\item Adopting private models in a double branch manner can nearly always lead to better performances than FedAvg~\cite{FedAvg}, while this is not true for Shakespeare.
		\item Taking single branch privatization ways almost always perform worse.
		\item The observations contradict the natural assumptions in Sect.~\ref{sect-issues}.
		\item Privatizing an encoder in a double branch manner, i.e., ``AaB'', can almost always lead to better results, which should be prioritized when faced with a novel FL non-iid scene. Following this, the ``ABb'' should be taken for a try.
	\end{itemize}
	
	\begin{figure}[htbp]
		\centering
		
		\begin{minipage}{0.45\linewidth}
			\centering
			\includegraphics[width=\linewidth]{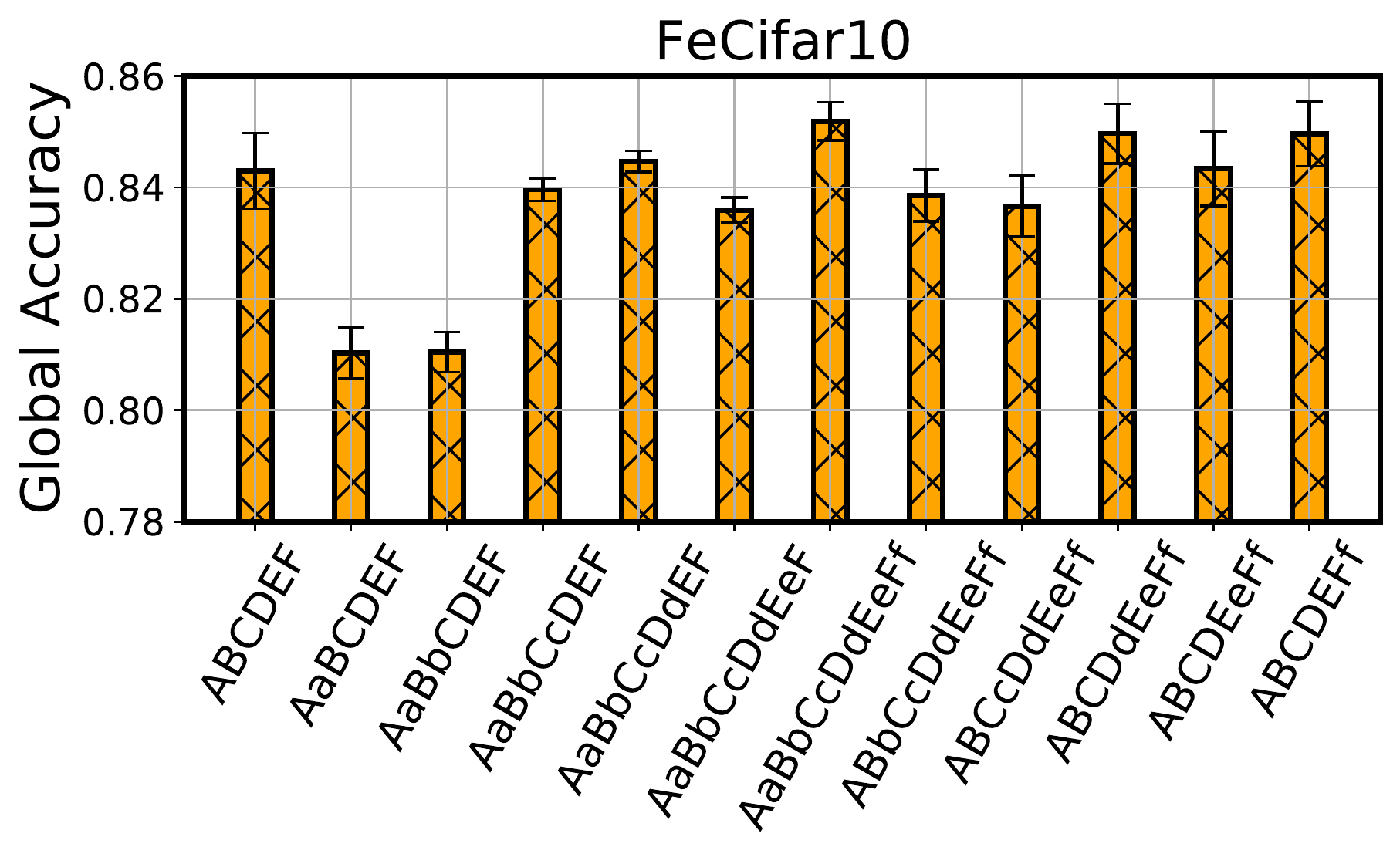}
			\centerline{(a) FeCifar10, Aggregation}
		\end{minipage}
		\quad
		\begin{minipage}{0.45\linewidth}
			\centering
			\includegraphics[width=\linewidth]{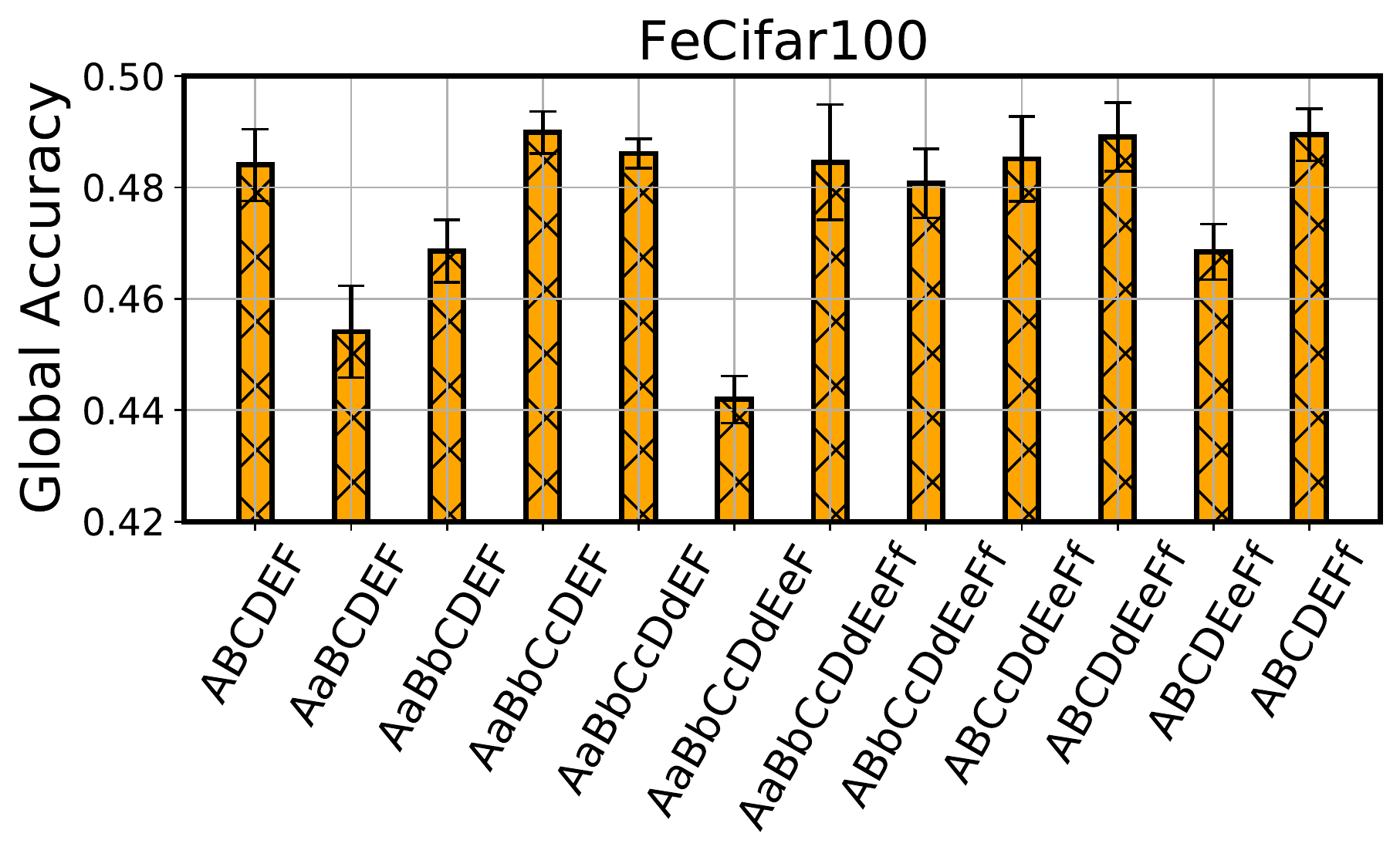}
			\centerline{(b) FeCifar100, Aggregation}
		\end{minipage}
		
		\begin{minipage}{\linewidth}
			\centering
			\includegraphics[width=\linewidth]{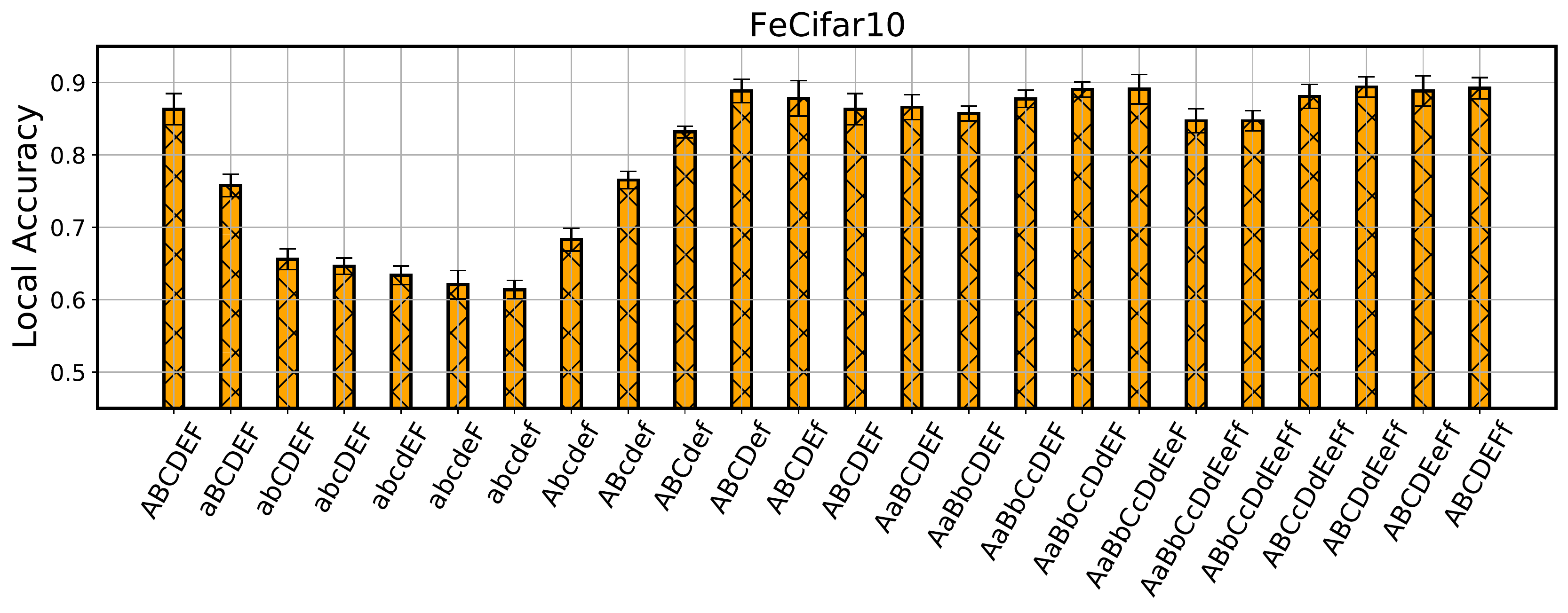}
			\centerline{(c) FeCifar10, Personalization}
		\end{minipage}
		
		\begin{minipage}{1.0\linewidth}
			\centering
			\includegraphics[width=\linewidth]{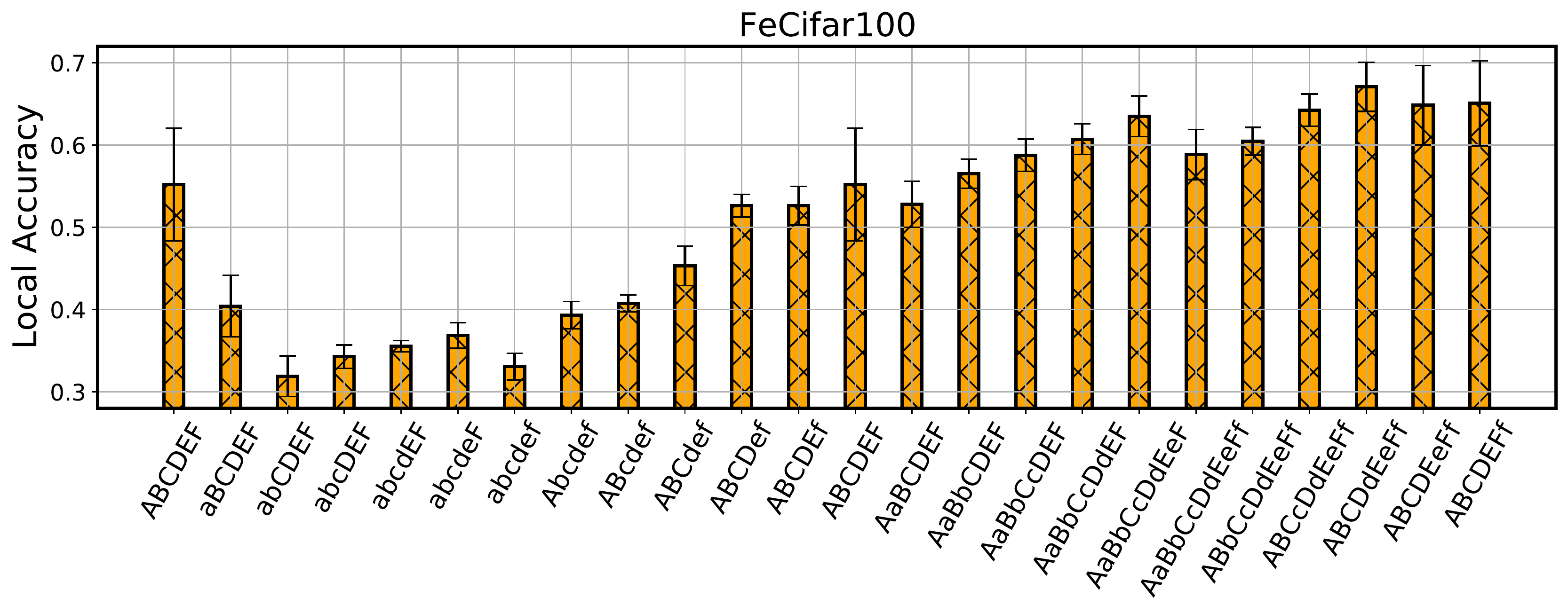}
			\centerline{(d) FeCifar100, Personalization}
		\end{minipage}
		
		\centering
		\caption{{\small Aggregation and personalization performances of different fine-grained privatization ways on FeCifar10 and FeCifar100. The top row shows the aggregation performances, while the bottom two rows show the personalization performances.}}
		\label{fig-fine-vgg11}
	\end{figure}

	Owing to the observation that none of the privatization ways performs better on Shakespeare, we investigate this scene with more settings, e.g., $Q=0.1$ and $E=5$. For $Q=0.1$, we increase the client selection ratio from $0.01$ to $0.1$, and investigate whether the privatization is relevant to the number of participating clients or not; for $E=5$, we want to find whether the performances of privatization ways are related to the local training steps or not. For each setting, we record the performances in Fig.~\ref{fig-shake-more}. We can still find that the privatization ways are useless. {\em Hence, we guess that Shakespeare may not be a severe non-iid scene that does not need any privatized components.}

	\subsubsection{Fine-grained Privatization} \label{sect-fine-privatization}
	Then we investigate the aggregation and personalization performances via fine-grained network splits. Specifically, we investigate the VGG11 on FeCifar10 and FeCifar100. The fine-grained network splits can be found in Fig.~\ref{fig-networks}. Here, we only report the average of the last 5 test rounds' performances and corresponding deviations. We plot the performances of various privatization ways in Fig.~\ref{fig-fine-vgg11}. Because the reported results only denote the performances when obtaining convergence, the observations will be a little different from the above analysis. The aggregation performances on FeCifar10 and FeCifar100 do not present any regular changes, while the obtained personalization performances are interesting. We can obviously observe that the personalization performances gradually decrease from ``ABCDEF'' to ``abcdeF'', or from ``ABCDEF'' to ``Abcedf'' in the single branch privatization ways. Individually training, i.e., ``abcdef'', obtains the worst personalization performances. {\em With more privatization blocks, no matter the base or the top layers, the performances degrade monotonically.} On FeCifar10 and FeCifar100, the double branch architectures can obtain similar personalization performances. {\em This again verifies that the single branch privatization ways will lead to worse performances, and we should take a double branch one.}
	
	\begin{figure}[htbp]
		\centering
		
		\begin{minipage}{0.45\linewidth}
			\centering
			\includegraphics[width=\linewidth]{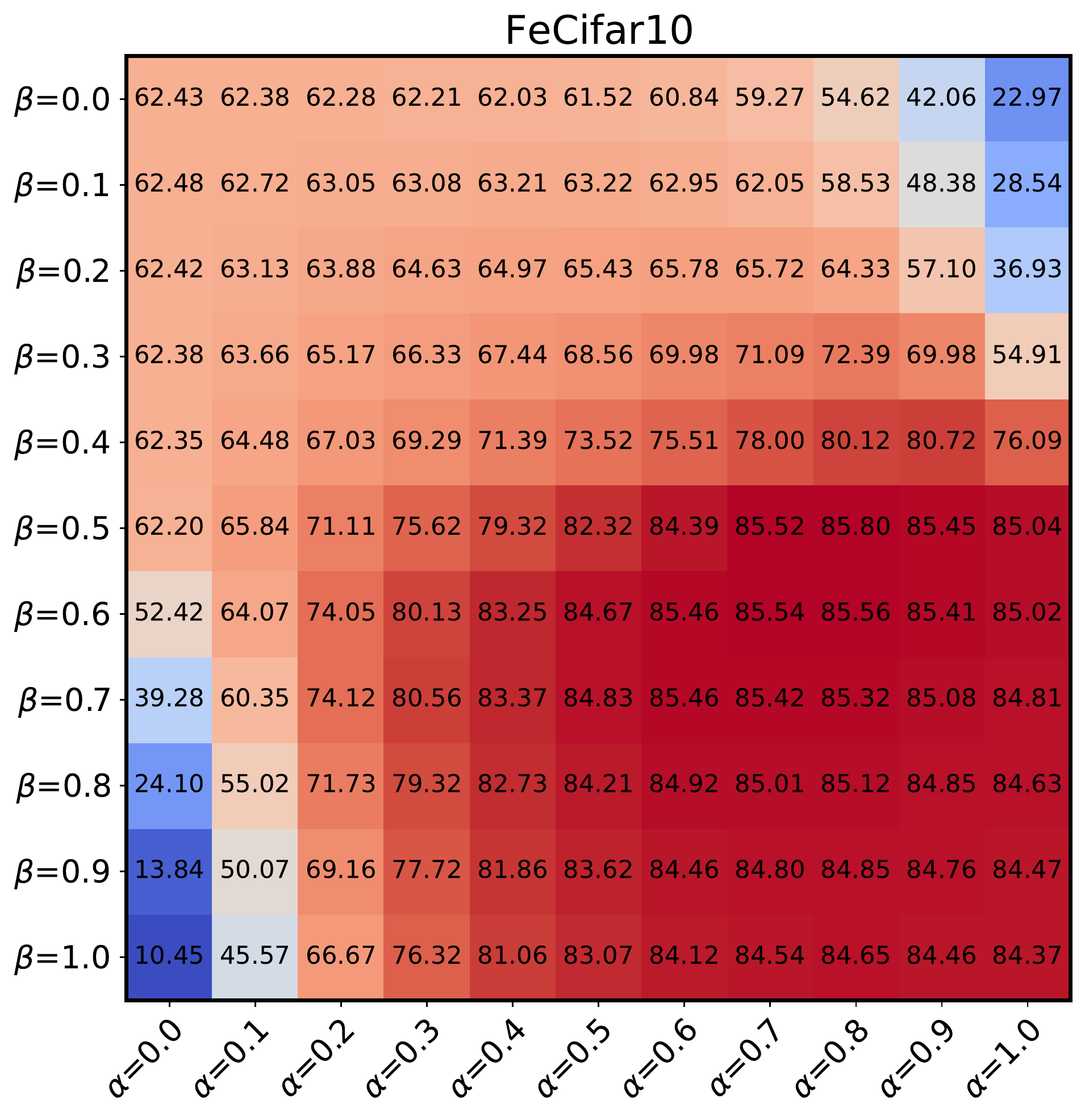}
			\centerline{(a) FeCifar10}
		\end{minipage}
		\quad
		\begin{minipage}{0.45\linewidth}
			\centering
			\includegraphics[width=\linewidth]{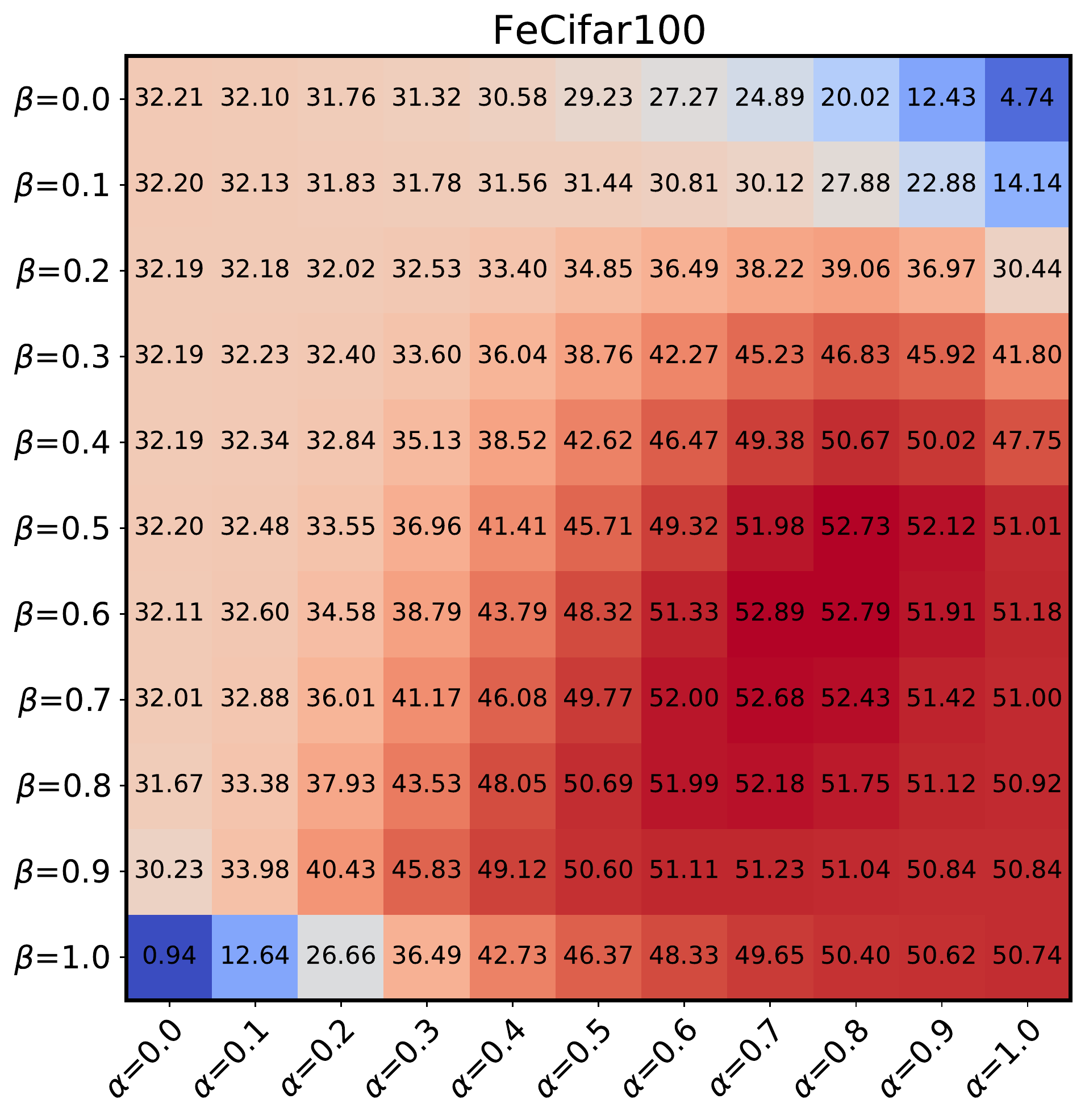}
			\centerline{(b) FeCifar100}
		\end{minipage}
		\begin{minipage}{0.45\linewidth}
			\centering
			\includegraphics[width=\linewidth]{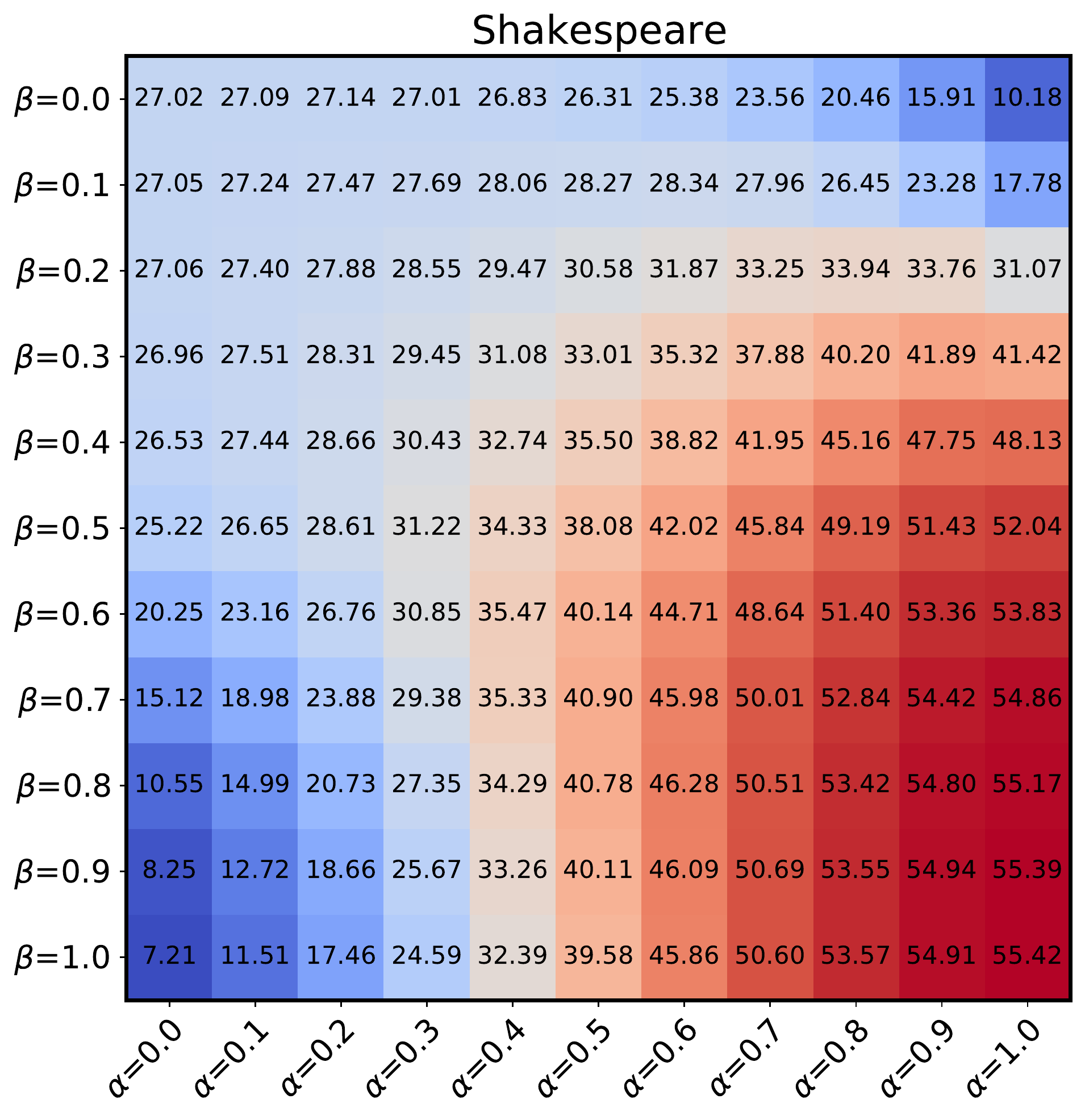}
			\centerline{(c) Shakespeare}
		\end{minipage}
		\quad
		\begin{minipage}{0.45\linewidth}
			\centering
			\includegraphics[width=\linewidth]{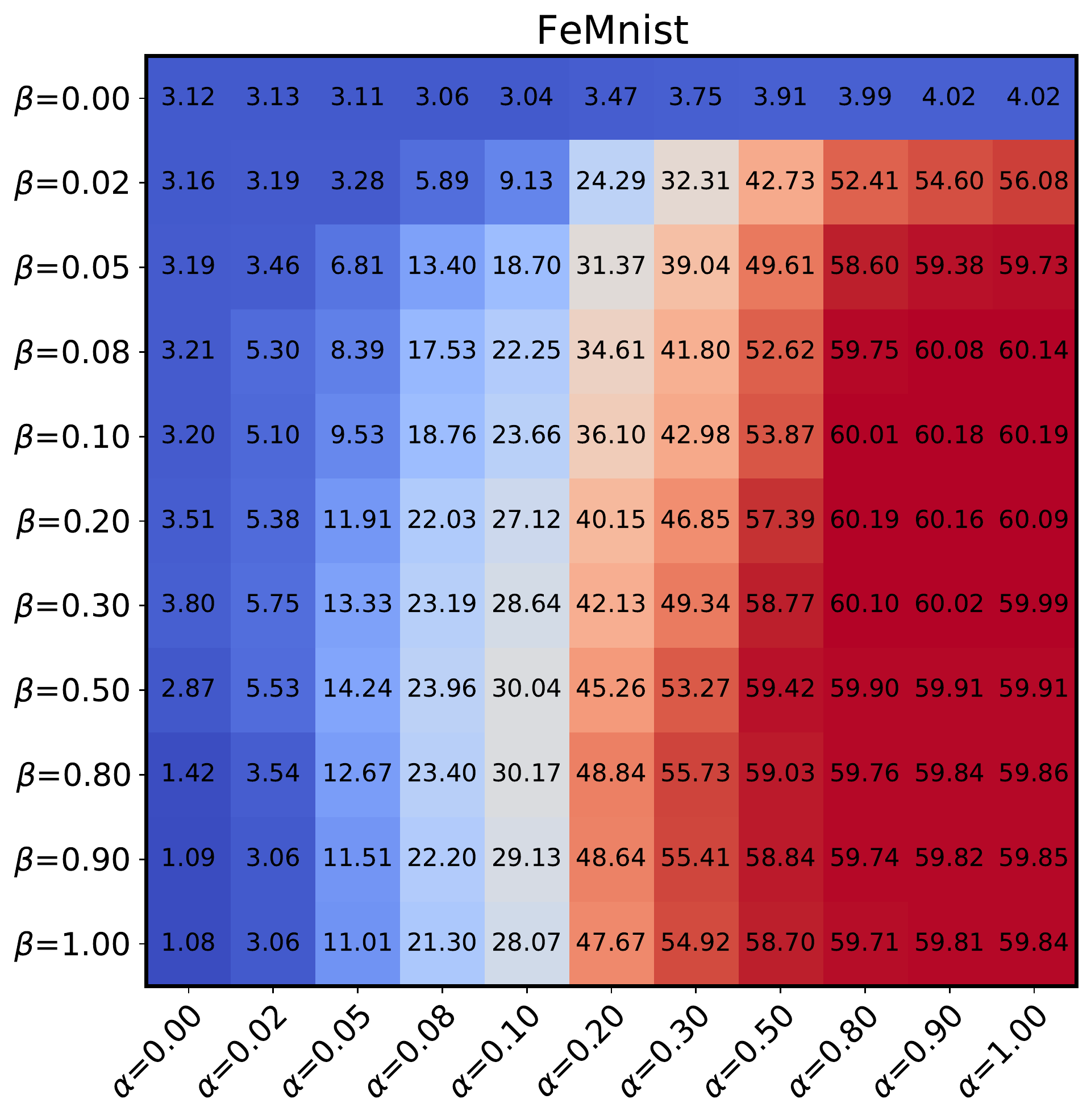}
			\centerline{(d) FeMnist}
		\end{minipage}
		
		\centering
		\caption{{\small Personalization performances of interpolations between shared and private components. Every figure shows one benchmark.}}
		\label{fig-per-inter}
	\end{figure}

	\begin{figure}[htbp]
		\centering
		
		\begin{minipage}{0.3\linewidth}
			\centering
			\includegraphics[width=\linewidth]{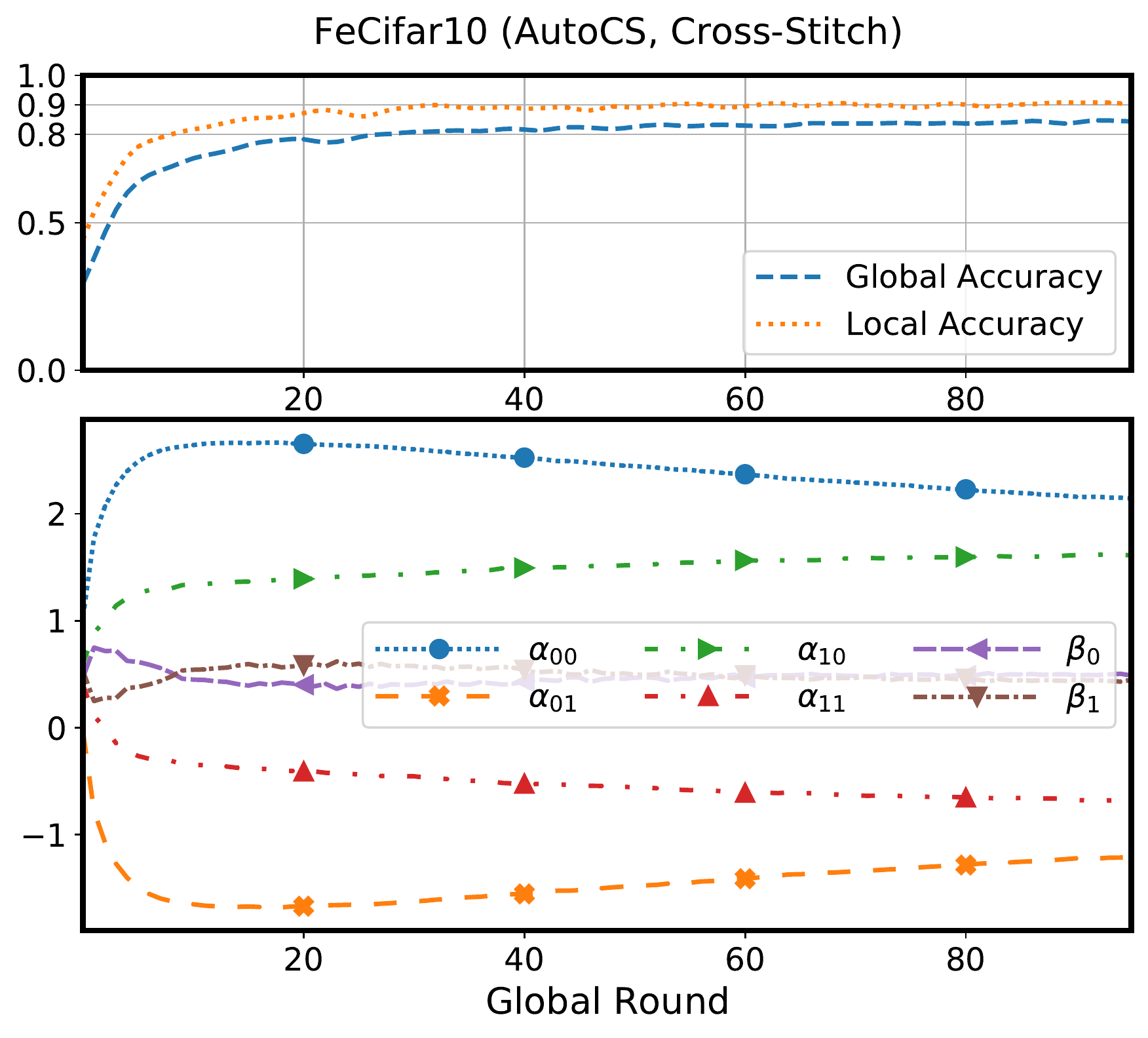}
			\centerline{(a) FeCifar10 AutoCS}
		\end{minipage}
		\quad
		\begin{minipage}{0.3\linewidth}
			\centering
			\includegraphics[width=\linewidth]{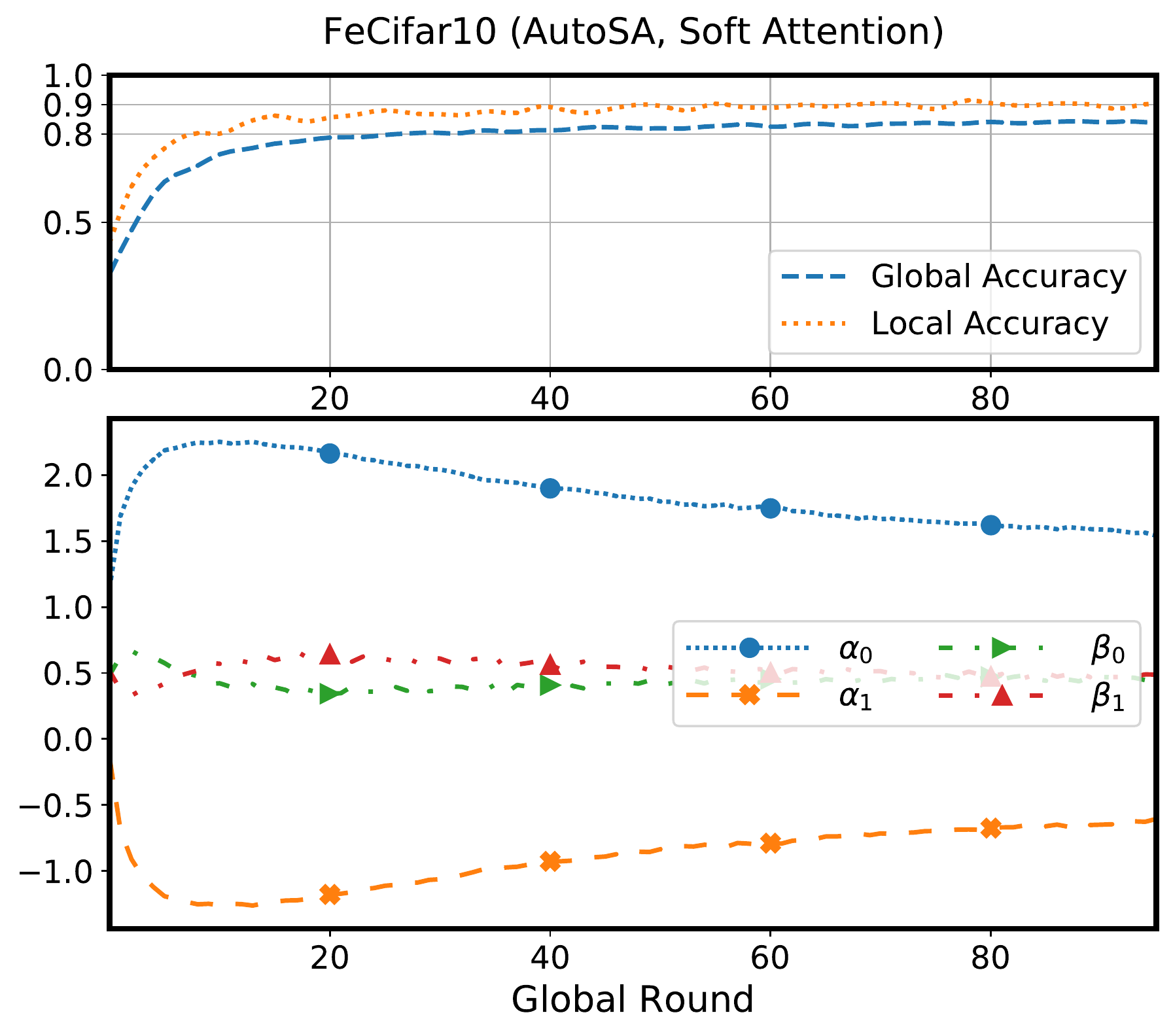}
			\centerline{(b) FeCifar10 AutoSA}
		\end{minipage}
		\quad
		\begin{minipage}{0.3\linewidth}
			\centering
			\includegraphics[width=\linewidth]{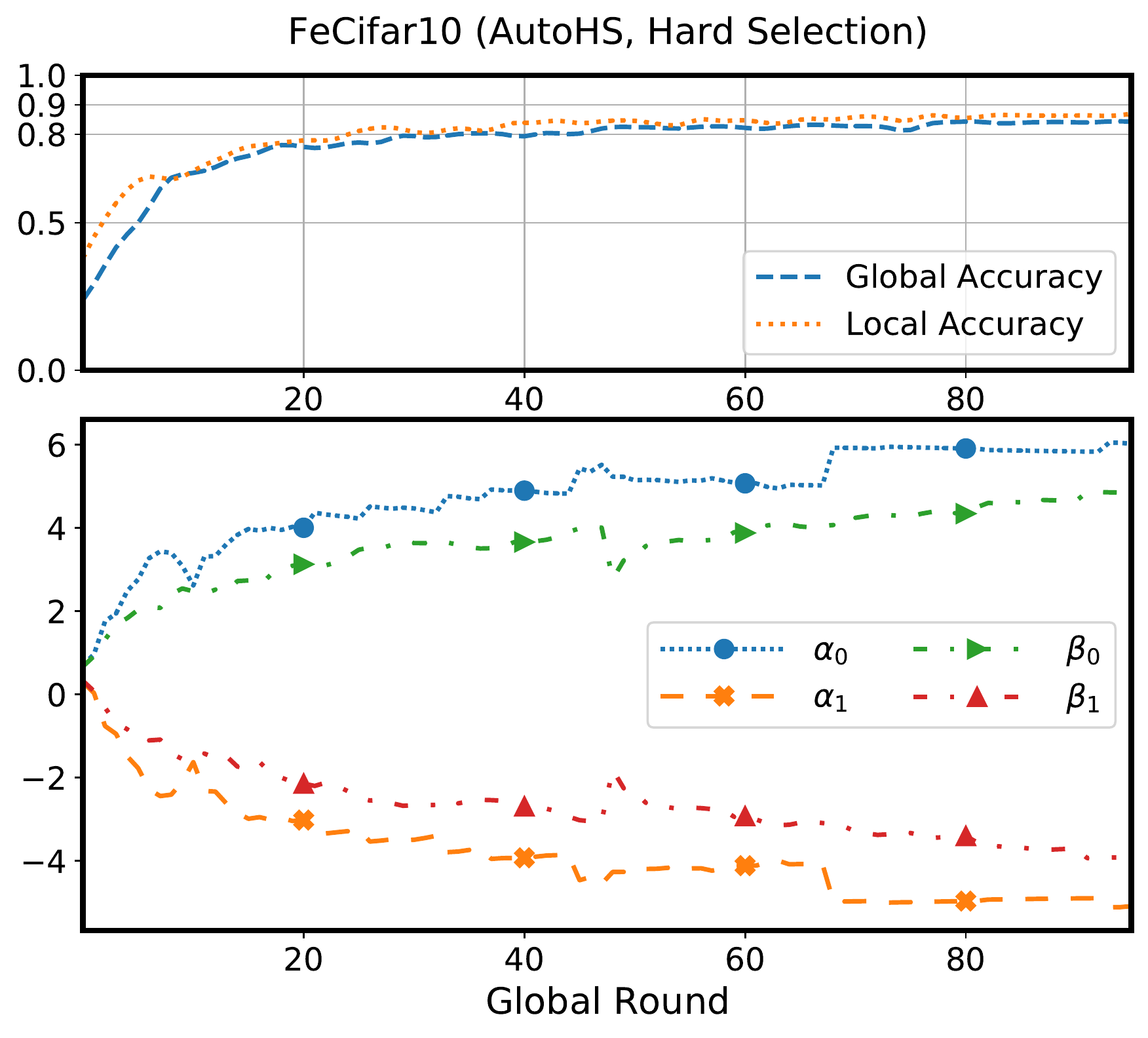}
			\centerline{(c) FeCifar10 AutoHS}
		\end{minipage}
		
		\begin{minipage}{0.3\linewidth}
			\centering
			\includegraphics[width=\linewidth]{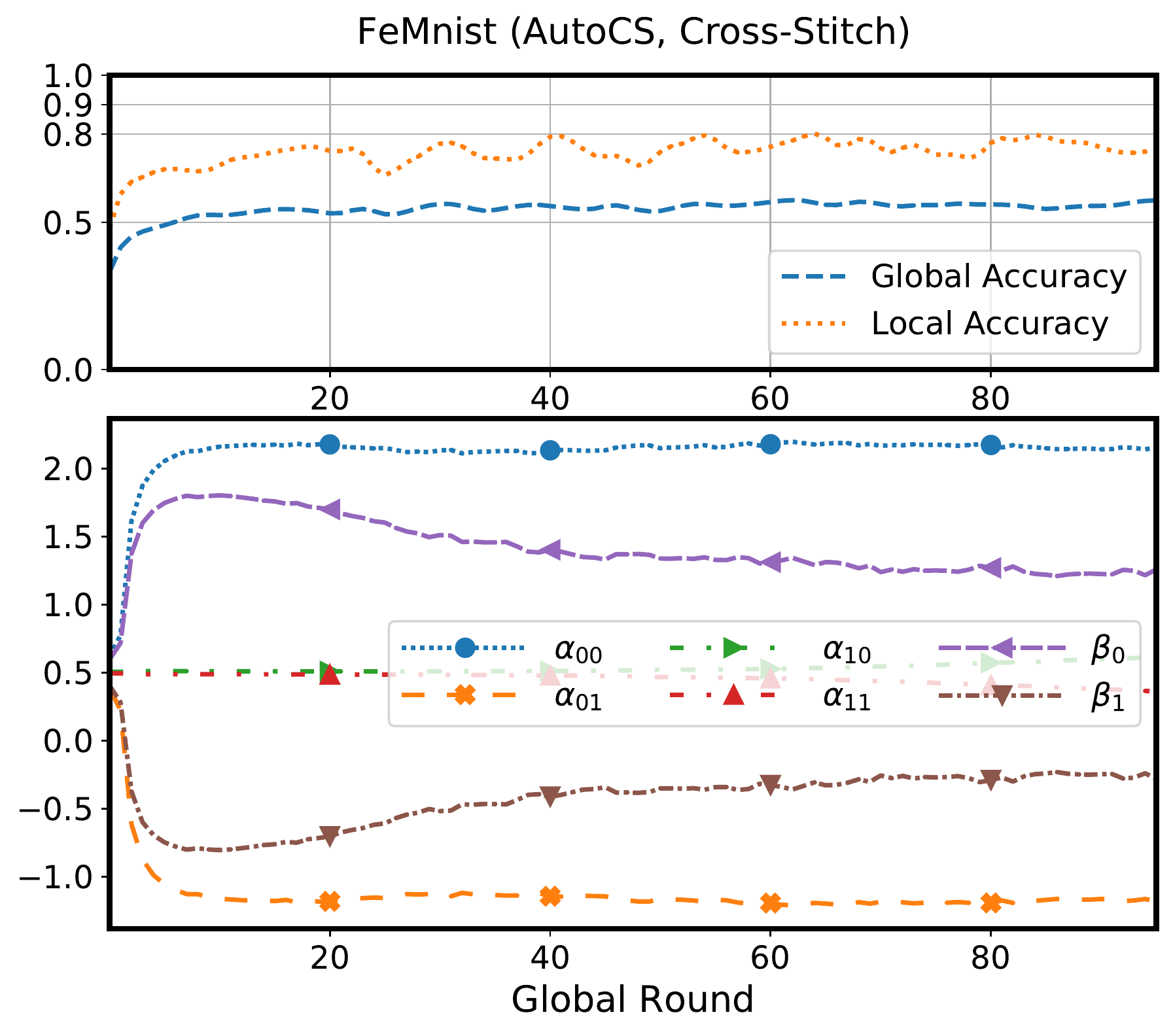}
			\centerline{(a) FeMnist AutoCS}
		\end{minipage}
		\quad
		\begin{minipage}{0.3\linewidth}
			\centering
			\includegraphics[width=\linewidth]{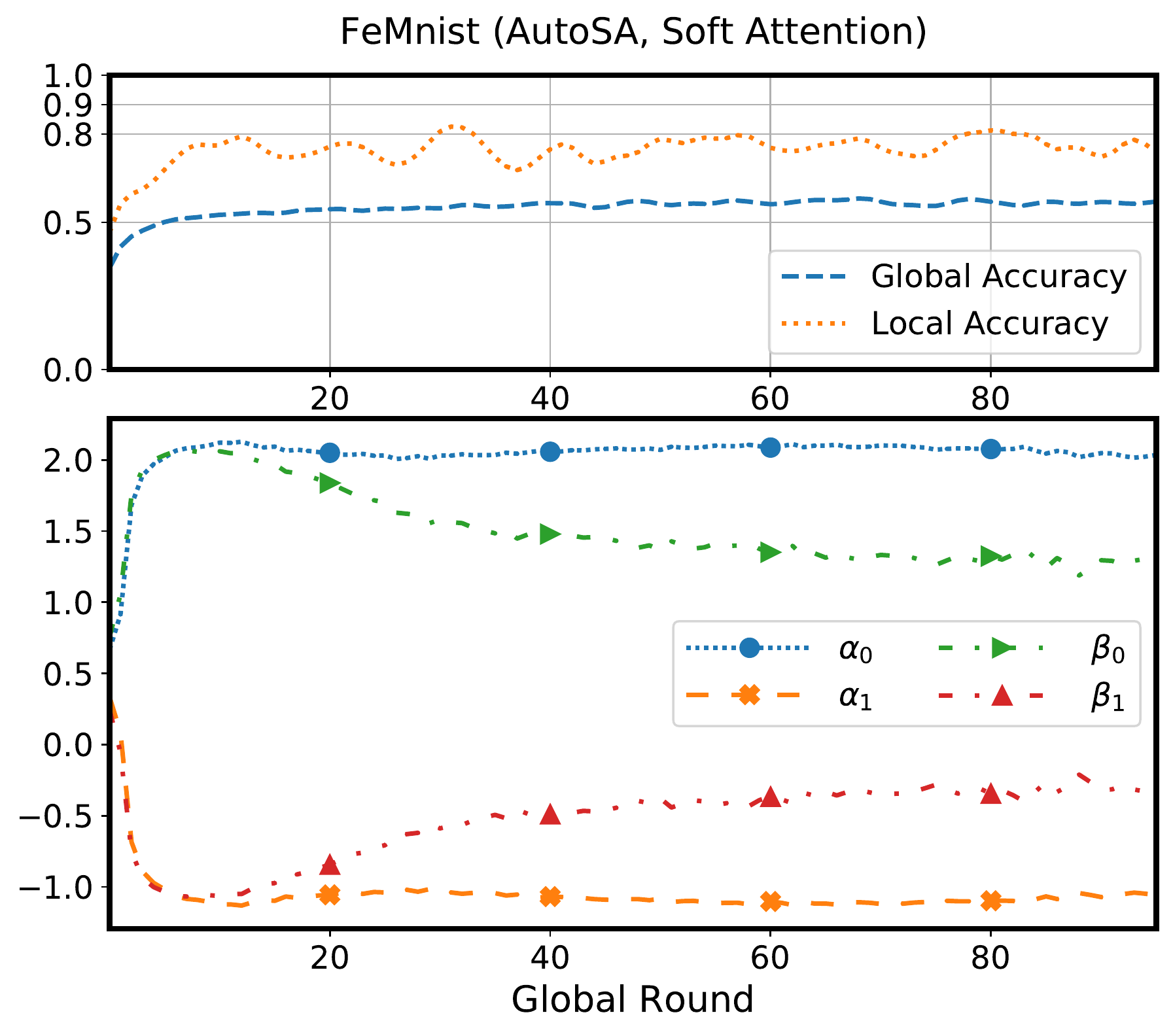}
			\centerline{(b) FeMnist AutoSA}
		\end{minipage}
		\quad
		\begin{minipage}{0.3\linewidth}
			\centering
			\includegraphics[width=\linewidth]{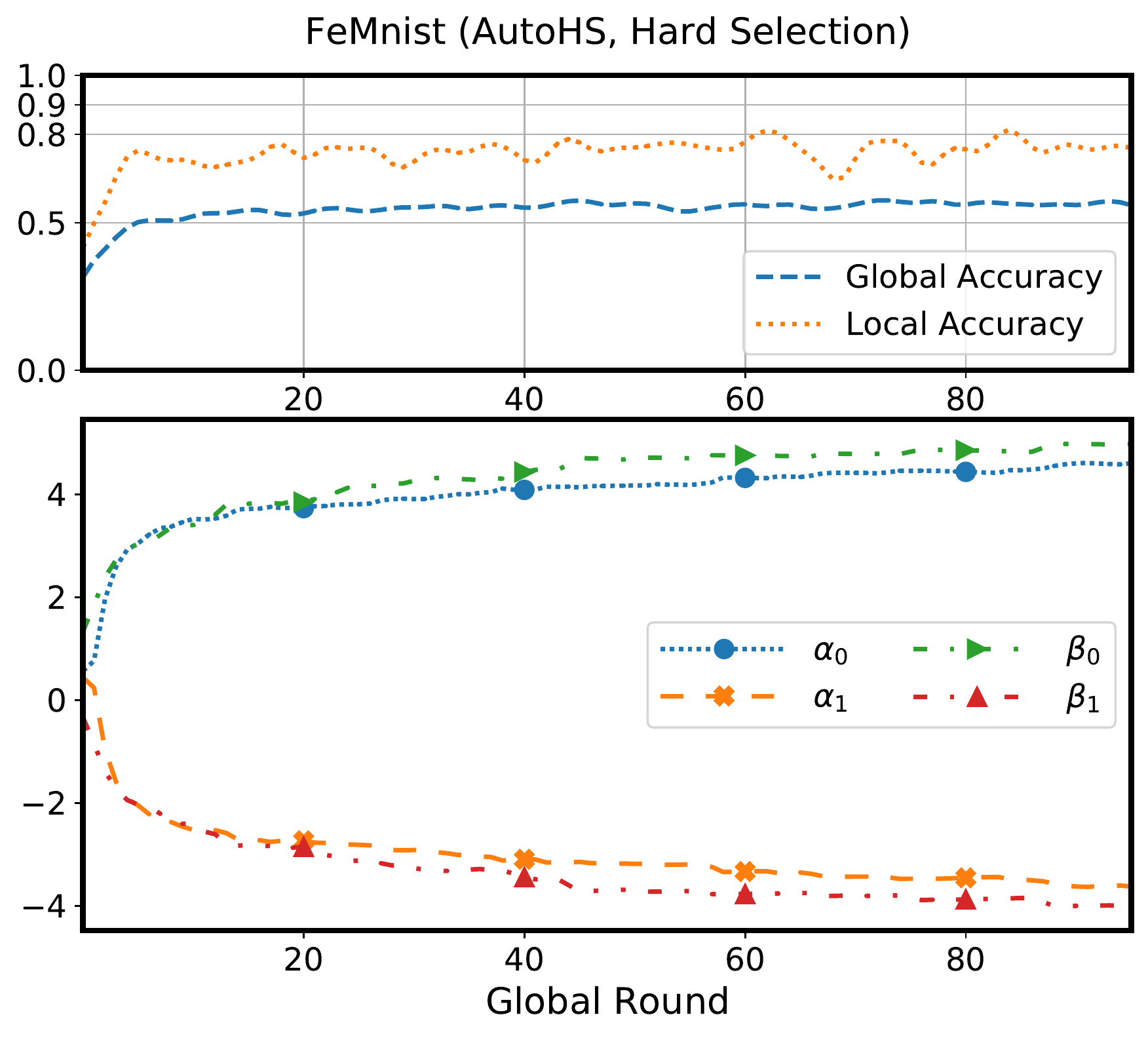}
			\centerline{(c) FeMnist AutoHS}
		\end{minipage}
		
		\centering
		\caption{{\small Aggregation performances, personalization performances, and the change of combination coefficients in the learning process of proposed three automatical methods. The two rows show the results on FeCifar10 and FeMnist, respectively. In each row, the three figures denote the three proposed algorithms, i.e., AutoCS, AutoSA, and AutoHS. In each figure, the top shows the aggregation and personalization performances, i.e., the global and local accuracy, the bottom shows the change of coefficients.}}
		\label{fig-auto-result}
	\end{figure}
		
	\subsection{Results of Automatical Algorithms}
	\label{sect-auto-learn}
	
	The above experimental studies investigate where to aggregate via enumerating possible privatization ways. In this section, we aim to automatically learn where to aggregate through a simple yet interesting interpolation trick and the proposed algorithms in Sect.~\ref{sect-algos}.
	
	We first take the architecture of ``AaBb'' to aggregate a global model and learn private models for each client. The global model contains a global encoder $E_{\ts}$ and a global classifier $C_{\ts}$. For each client, it owns a local encoder $E_{\tp}^k$ and a local classifier $C_{\tp}^k$. We investigate the interpolations of shared and private features, shared and private outputs. Specifically, for each client, we can obtain the shared features $\bfh_{\ts}^k$ and private features $\bfh_{\tp}^k$ as in Eq.~\ref{eq-feat}. We omit the index of samples for simplification. Then we get an interpolation of the features via $\hat{\bfh}=\alpha \bfh_{\ts}^k + (1-\alpha)\bfh_{\tp}^k$. Next, we obtain predictions via feeding $\hat{\bfh}$ into both $C_{\ts}$ and $C_{\tp}^k$, and we denote the probabilities (i.e., predictions after softmax) as $\bfp_{\ts}^k$ and $\bfp_{\tp}^k$. We also take an interpolation via $\hat{\bfp}=\beta \bfp_{\ts}^k + (1-\beta) \bfp_{\tp}^k$. Finally, we calculate the personalization accuracy via the predicted $\hat{\bfp}$ and the real labels. Varying $\alpha \in [0,1]$ and $\beta \in [0,1]$, we can obtain the performances of different privatization ways in a post-processing manner. For example, if $\alpha=1.0$ and $\beta=1.0$, it degenerates into the ``AB'' way. We plot the personalization results in Fig.~\ref{fig-per-inter}, and the results are evaluated as in Def.~\ref{def-personalization}. For FeMnist, we take a non-uniform change of $\alpha$ and $\beta$ to better show the change of results. For FeCifar10, the best interpolation lies around $\alpha=0.8$ and $\beta=0.5$; for FeCifar100, the best one lies around $\alpha=0.7$ and $\beta=0.6$. Hence, these two scenes show that taking both the combinations of encoders and classifiers is preferred, which conforms to the results in Sect.~\ref{sect-coarse-privatization} that fusing features (i.e., ``AaB'') or fusing outputs (i.e., ``ABb'') could improve the FL performances. For Shakespeare, setting $\alpha=1.0$ and $\beta=1.0$ can obtain the best results, which again verifies that taking no private components on Shakespeare is more suitable. For FeMnist, setting $\alpha=1.0$ and $\beta=0.1$ is better, implying that a shared encoder and an interpolated classifier will be more suitable, i.e., an architecture similar to ``ABb''. This also conforms to the performances of ``ABb'' in Fig.~\ref{fig-coarse-agg} (d) and Fig.~\ref{fig-coarse-per} (d).
	
	Additionally, we can find that the changes are regular, and the tendencies along with the change of $\alpha$ and $\beta$ are apparent. This could be an easy procedure to investigate where to aggregate when faced with a novel non-iid FL scene. Namely, we can first take the ``AaBb'' architecture to train both the global model and personalized models preliminarily, and then we use this interpolation trick to find which combination performs better. Finally, we can use the architecture of the mined one for regular training and application. Take Shakespeare as an example, we can find that setting $\alpha=1.0$ and $\beta=1.0$ could lead to better results, and hence we could take the ``AB'' way, which {\em saves vast training cost of enumerating all possible privatization ways}. From another aspect, this trick could possibly serve as an efficient way to measure whether a scene is a severe non-iid scene that needs private components to improve performances.
	
	Next, we report the performances and the change of coefficients in $\psi$ with the three proposed algorithms, i.e., AutoCS, AutoSA, and AutoHS introduced in  Sect.~\ref{sect-algos}. The results of FeCifar10 and FeMnist are shown in Fig.~\ref{fig-auto-result}. In each figure, the top presents the convergence curves of both aggregation and personalization, i.e., the global and local accuracy. Compared with previous results, we can find that the algorithms can obtain comparable performances with the best performances in Fig.~\ref{fig-coarse-agg} and Fig.~\ref{fig-coarse-per}, which implies that the designed algorithms can effectively learn where to aggregate. For comparisons of the three algorithms, AutoHS obtains a little worse personalization performance in FeCifar10, which may be related to the Gumbel sampling's randomness. Then we have a deep look at the learned coefficients in the FL process. The illustrations of these coefficients can be found in Fig.~\ref{fig-auto-learn}. In AutoCS, we find that $\alpha_{0,0}$ tends to be larger than $\alpha_{0,1}$ on both FeCifar10 and FeMnist, which implies that shared classifier is more inclined to use the features extracted by the shared encoder. However, the relation of $\beta_0$ and $\beta_1$ differs significantly on FeCifar10 and FeMnist, where the former tends to combine the shared and private classifiers and the latter assigns a larger weight to the shared classifier. For AutoSA, similar phenomena can also be observed. In AutoHS, the $\alpha_0$ and $\beta_0$ tend to be larger than $\alpha_1$ and $\beta_1$, which assigns larger weights to the shared components.
	Noting that the smaller coefficients will not be completely suppressed to zeroes after softmax, e.g., $\text{SoftMax(2.0/2.0, -1.0/2.0)}$ returns 0.82 and 0.18 correspondingly, and hence the learned architecture still implies a soft combination among shared and private components. The coefficients may be related to the balancing factor $\alpha$ in Eq.~\ref{eq:ps-loss}. We will explore this in future work.
	
	The proposed interpolation trick and automatical algorithms are efficient to implement and could be taken to search for an appropriate privatization way preliminarily when given a novel non-iid scene. Some other ways could also be explored, e.g., interpolating the parameters instead of the features and outputs, learning combination weights locally, or assigning different combination weights for different neurons. These possible studies are also left as future work.

	\section{Conclusion}
	Motivated by the observation that existing FL methods take various privatization ways in the learning process, we aimed to explore whether the privatization ways have any potential relations to the types of non-iid scenes.
	We selected four FL benchmarks containing both covariate shift and label shift non-iid scenes, designed coarse-grained and fine-grained privatization ways, and took abundant experimental studies to analyze both aggregation and personalization performances.
	We observed several exciting phenomena, e.g., privatizing the encoder can lead to better performances even in label shift non-iid scenes; none of the investigated privatization ways perform better on Shakespeare; the double branch privatization ways perform better when compared to the single branch ones.
	These observations suggest that privatizing an encoder in a double branch manner (i.e., ``AaB'') should be prioritized for a novel non-iid scene, regardless of which type of data heterogeneity, and the next to adopt is privatizing an additional classifier (i.e., ``ABb'').
	We also provided efficient ways to automatically learn where to aggregate via an interpolation trick and three proposed automatical algorithms. The results verify that these proposed methods could serve as a preliminary try to explore which layers to privatize.
	To be brief, our investigation provides insights for further studies of where to aggregate or privatize in FL, e.g., applying advanced neural architecture search techniques to FL and quantifying the non-iid level of faced scenes.

	\bibliographystyle{ACM-Reference-Format}
	\bibliography{fedps}
	

\end{document}